\documentclass[sigconf]{aamas}

\usepackage{balance} % for balancing columns on the final page

\usepackage[ruled,vlined]{algorithm2e}
\usepackage{caption}
\usepackage{subcaption}
\usepackage{float}
\usepackage{algpseudocode}
\usepackage{amsmath}
\usepackage{multirow}
\usepackage{booktabs}
\usepackage{appendix}
\newcommand{\pms}[1]{\scriptsize{$\pm$#1}}
\DeclareFontFamily{U}{stix2bb}{}
\DeclareFontShape{U}{stix2bb}{m}{n} {<-> stix2-mathbb}{}

\providecommand{\doi}[1]{\acmDOI{#1}}
\doi{WSBG8309}

\makeatletter
\gdef\@copyrightpermission{
  \begin{minipage}{0.2\columnwidth}
   \href{https://creativecommons.org/licenses/by/4.0/}{\includegraphics[width=0.90\textwidth]{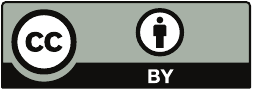}}
  \end{minipage}\hfill
  \begin{minipage}{0.8\columnwidth}
   \href{https://creativecommons.org/licenses/by/4.0/}{This work is licensed under a Creative Commons Attribution International 4.0 License.}
  \end{minipage}
  \vspace{5pt}
}
\makeatother

\setcopyright{ifaamas}
\acmConference[AAMAS '26]{Proc.\@ of the 25th International Conference
on Autonomous Agents and Multiagent Systems (AAMAS 2026)}{May 25 -- 29, 2026}
{Paphos, Cyprus}{C.~Amato, L.~Dennis, V.~Mascardi, J.~Thangarajah (eds.)}
\copyrightyear{2026}
\acmYear{2026}
\acmDOI{}
\acmPrice{}
\acmISBN{}

\acmSubmissionID{1797}

\title[AAMAS-2026 Formatting Instructions]{TAAM:Inductive Graph-Class Incremental Learning with Task-Aware Adaptive Modulation}

\author{Jingtao Liu}
\affiliation{
  \institution{University of Science and Technology of China}
  \city{HeFei}
  \country{China}}
\email{jingt-liu@mail.ustc.edu.cn}

\author{Xinming Zhang}
\authornote{Corresponding author.}
\affiliation{
  \institution{University of Science and Technology of China}
  \city{HeFei}
  \country{China}}
\email{xinming@ustc.edu.cn}

\begin{abstract}
Graph Continual Learning (GCL) aims to solve the challenges of streaming graph data.  However, current methods often depend on replay-based strategies, which raise concerns like memory limits and privacy issues, while also struggling to resolve the stability-plasticity dilemma.  In this paper, we suggest that lightweight, task-specific modules can effectively guide the reasoning process of a fixed GNN backbone.  Based on this idea, we propose \textbf{T}ask-\textbf{A}ware \textbf{A}daptive \textbf{M}odulation (TAAM).  The key component of TAAM is its lightweight Neural Synapse Modulators (NSMs).  For each new task, a dedicated NSM is trained and then frozen, acting as an “expert module.”  These modules perform detailed, node-attentive adaptive modulation on the computational flow of a shared GNN backbone.  This setup ensures that new knowledge is kept within compact, task-specific modules, naturally preventing catastrophic forgetting without using any data replay.  Additionally, to address the important challenge of unknown task IDs in real-world scenarios, we propose and theoretically prove a novel method named Anchored Multi-hop Propagation (AMP).  Notably, we find that existing GCL benchmarks have flaws that can cause data leakage and biased evaluations.  Therefore, we conduct all experiments in a more rigorous inductive learning scenario.  Extensive experiments show that TAAM comprehensively outperforms state-of-the-art methods across eight datasets.Code and Datasets are available at: \url{https://github.com/1iuJT/TAAM_AAMAS2026}.
\end{abstract}



\newcommand{\BibTeX}{\rm B\kern-.05em{\sc i\kern-.025em b}\kern-.08em\TeX}

\begin{document}

% %%% The following commands remove the headers in your paper. For final 
% %%% papers, these will be inserted during the pagination process.

\pagestyle{fancy}
\fancyhead{}

% %%% The next command prints the information defined in the preamble.

\maketitle 

% %%%%%%%%%%%%%%%%%%%%%%%%%%%%%%%%%%%%%%%%%%%%%%%%%%%%%%%%%%%%%%%%%%%%%%%%

% INTRO: 7/10
\section{Introduction}
\label{1.intro}

%虽然图神经网络（gnn）在静态图上取得了显著的成功，但现实世界的图数据通常是动态的和不断发展的。为了解决这个问题，持续图学习（CGL）已经成为一个关键的研究领域\cite{Tian2024ContinualLO}，其中模型必须从不断发展的任务流中学习，同时保留先前获得的知识（图\ref{fig:single_column}，顶部）。在图形类增量学习范式中，图形类增量学习（GCIL）是一个特别具有挑战性和实用性的前沿。它的核心原则是任务标识符（id）在推理过程中不可用，这一场景反映了现实世界的应用程序，如药物发现，其中出现了新的化合物族\cite{Zhang2022CGLBBT}，或者社交网络，其中用户社区随着时间的推移而发展\cite{Wang2020LifelongGL}。这种与任务无关的要求带来了双重挑战：首先，克服灾难性遗忘，其次，解决“任务间类分离”问题——在没有明确上下文的情况下，很难从不同的任务中区分相似的类\cite{Su2024TowardsRG}。

\begin{figure}[t]
    \centering
    \Description{A two-part diagram. The top illustrates a continuous stream of graph data with new nodes and tasks. The bottom shows a bar chart comparing performance, highlighting the stability-plasticity dilemma where replay methods suffer from reduced plasticity.}
    \includegraphics[width=1\columnwidth]{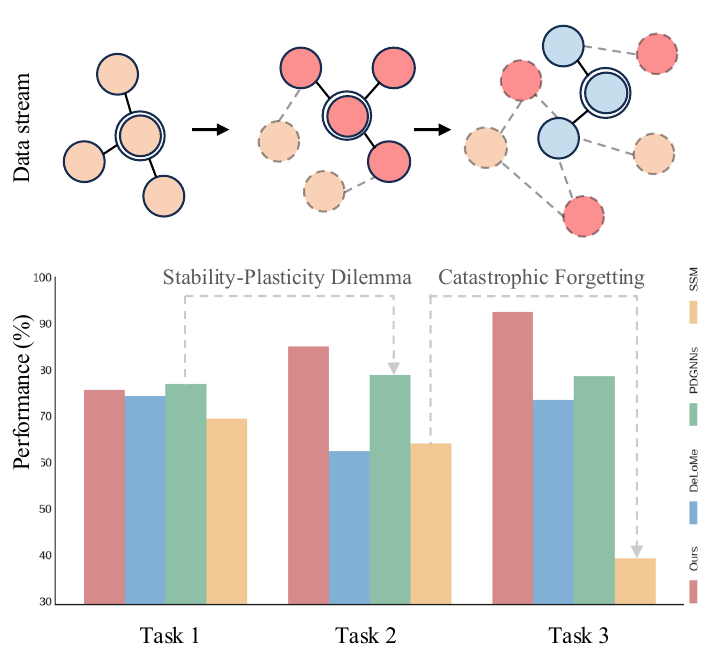}
    \caption{Top: A typical CGL scenario where a model must adapt to a stream of new tasks (indicated by new node colors) in an evolving graph. Bottom: An illustration of the Stability-Plasticity Dilemma on the Citeseer benchmark. While strong replay-based methods effectively mitigate the catastrophic forgetting of early tasks, they exhibit diminished plasticity.}
    \label{fig:single_column}
\end{figure}

While Graph Neural Networks (GNNs) have demonstrated remarkable success on static graphs, real-world graph data is often dynamic and evolving.To address this, Continual Graph Learning (CGL) has become a key research field~\cite{Tian2024ContinualLO}, where models must learn from an evolving stream of tasks while retaining previously acquired knowledge (Figure~\ref{fig:single_column}, top). Among CGL paradigms, Graph Class-Incremental Learning (GCIL) presents a particularly challenging and practical frontier. Its core tenet is that task identifiers (IDs) are unavailable during inference, a scenario mirroring real-world applications like drug discovery, where new compound families emerge~\cite{Zhang2022CGLBBT}, or social networks, where user communities evolve over time~\cite{Wang2020LifelongGL}. This task-agnostic requirement introduces a dual challenge: first, overcoming catastrophic forgetting, and second, resolving the ``inter-task class separation'' problem---the difficulty of distinguishing between similar classes from different tasks without explicit context~\cite{Su2024TowardsRG}.

%为了应对这些挑战，研究人员主要探索了三种范式：基于正则化、基于重播和参数隔离方法\cite{Zhang2022CGLBBT,Zhang2023ContinualLO}。基于重播的方法存储和预演历史数据，可以有效地减轻遗忘，但会带来巨大的计算和存储开销，以及潜在的数据隐私风险。关键的是，他们面临着保存旧知识（稳定性）和获取新信息（可塑性）之间的基本权衡（图\ref{fig:single_column}，底部）。相比之下，参数隔离方法为克服遗忘提供了更直接的途径。通过将不同的参数用于不同的任务，这些方法从结构上防止了知识干扰\cite{Zhang2023ContinualLO,Zhang2021HierarchicalPN}。

To address these challenges, researchers have primarily explored three paradigms: regularization-based, replay-based, and parameter-isolation methods~\cite{Zhang2022CGLBBT,Zhang2023ContinualLO}. Replay-based approaches, which store and rehearse historical data, can effectively mitigate forgetting but introduce significant computational and storage overhead, along with potential data privacy risks. Critically, they face a fundamental trade-off between preserving old knowledge (stability) and acquiring new information (plasticity)(Figure~\ref{fig:single_column}, Bottom).In contrast, parameter-isolation methods offer a more direct path to overcoming forgetting. By dedicating distinct parameters to different tasks, these approaches structurally prevent knowledge interference~\cite{Zhang2023ContinualLO,Zhang2021HierarchicalPN}.

%最近，提示学习（Prompt Learning）作为一种参数高效的实现方法，在计算机视觉领域取得了巨大的成功\cite{Zhang2022CGLBBT,Zhang2023ContinualLO}。这种范式很快被适应到图域\cite{Liu2023GraphPromptUP,Wang2025PromptDrivenCG}，像TPP这样的开创性工作推断任务id，用小的、可学习的“图提示”\cite{niu2024replayandforgetfree}来调节冻结的GNN。然而，这种模式的成功是由一项重大成本支撑的——“训练前税”。其有效性关键取决于一个强大的、预先训练的GNN主干，以确保知识可转移\cite{Fang2022UniversalPT,Liu2023GapformerGT}。此外，大多数图形提示方法采用单一的通用提示来调节模型的输入或输出，仅提供对GNN内部计算流的粗粒度控制，从而限制了其表达和自适应能力\cite{niu2024replayandforgetfree}。这种对预训练的依赖在图域尤为繁重。与NLP和视觉不同，缺乏标准化、通用的预训练gnn，跨异构图结构对齐数据仍然是一个巨大的挑战\cite{Hu2020OpenGB,Tang2023GraphGPTGI}。

Recently, Prompt Learning has emerged as a highly effective parameter-efficient fine-tuning strategy, achieving great success in computer vision~\cite{Zhang2022CGLBBT,Zhang2023ContinualLO}. This paradigm was quickly adapted to the graph domain~\cite{Liu2023GraphPromptUP,Wang2025PromptDrivenCG}, with pioneering works like TPP conditioning a frozen GNN with small, learnable "graph prompts" based on inferred task IDs~\cite{niu2024replayandforgetfree}.However, the effectiveness of this paradigm is often predicated on the availability of a powerful, pre-trained GNN backbone to ensure robust knowledge transferability~\cite{Fang2022UniversalPT,Liu2023GapformerGT}. Moreover, many graph prompting methods employ universal prompts that offer only coarse-grained control over the GNN's internal computational flow, thus limiting their adaptive capacity~\cite{niu2024replayandforgetfree}. This reliance on pre-training presents unique challenges in the graph domain. Unlike in NLP and Computer vision standardized, universal pre-trained GNNs are not yet readily available, and aligning knowledge across heterogeneous graph structures remains a formidable task~\cite{Hu2020OpenGB,Tang2023GraphGPTGI}.
This confluence of challenges motivates a fundamental research question: Is it possible to design a CGL approach  that eliminates the dependency on a heavily pre-trained model, yet still achieves high performance in a efficient, replay-free manner?

In this paper, we tackle this challenge by rethinking how the stability-plasticity trade-off can be managed in a resource-efficient, replay-free context. Our approach maintains stability by leveraging a frozen GNN backbone and encapsulating task-specific knowledge within lightweight modulators. We argue that the key to unlocking plasticity lies in designing highly expressive, task-specific modules and, critically, a robust mechanism to identify and deploy the correct knowledge for a given task at inference time.

To this end, we introduce \textbf{T}ask-\textbf{A}ware \textbf{A}daptive \textbf{M}odulation (TAAM), a novel approach that operationalizes this principle. The architecture of TAAM is composed of two core innovations:

First, for plasticity, we design dynamic and lightweight \textbf{N}eural \textbf{S}ynapse \textbf{M}odulators (NSMs). For each task, a new NSM is created and trained to perform fine-grained, node-attentive modulation on the internal representations of the frozen GNN backbone. These modules act as task-specific experts, effectively steering the GNN's computational flow without altering its stable, foundational parameters.

Second, for stability, we address the critical challenge of task-ID inference. We propose and theoretically ground a novel method named \textbf{A}nchored \textbf{M}ulti-hop \textbf{P}ropagation (AMP) to generate highly robust and discriminative task prototypes.  For inference, AMP allows TAAM to accurately identify the task affiliation of incoming data, ensuring the correct expert NSM is selected. This precise retrieval mechanism is fundamental to preventing catastrophic forgetting. The modulated output is then processed by a unified classifier whose output dimension expands for new classes, preserving knowledge at the decision layer.We summarize our main contributions as follows.
\begin{itemize}
    \item We propose \textbf{TAAM}, a novel CGL approach that operates without reliance on data replay or extensive model pre-training, offering a truly resource-efficient solution.
    \item We introduce and provide theoretical grounding for Anchored Multi-hop Propagation (AMP), a robust method for task-ID inference that ensures model stability and prevents forgetting, validated across eight datasets.
    \item We propose a lightweight, and node-attentive modulator (NSM) that effectively incorporates task-specific knowledge and provides more expressive, fine-grained control over GNN representations than traditional static prompting methods.
    \item We identify limitations in existing GCL benchmarks and are the first to validate our approach in the more challenging and realistic inductive graph continual learning scenario, demonstrating its practical effectiveness.
\end{itemize}

% Related Works: 6/10
\section{Related Works}
\label{2.related_work}
% \vspace{-0.25em}

\textbf{Replay-based Methods.} A dominant paradigm in Continual Graph Learning (CGL) is the use of replay-based methods, which maintain a memory buffer of representative exemplars from past tasks. During training on a new task, these stored exemplars are rehearsed alongside the new data, mitigating catastrophic forgetting through joint optimization~\cite{liu2023cat, liu2024puma, Zhang2022SparsifiedSubgraphMemoryforCGRL, Zhang2024Topology-awareembeddingmemoryforcl}. The sophistication of what is stored has evolved, from representative nodes~\cite{Zhou2021Overcomingcatastrophicforgetting} to more complex structures. For instance, CaT~\cite{liu2023cat} and PUMA~\cite{liu2024puma} focus on graph condensation techniques to create small, synthetic graphs for replay. Others compress topological information, either by storing sparsified computation subgraphs (SSM~\cite{Zhang2022SparsifiedSubgraphMemoryforCGRL}) or by decoupling the model from the graph via compressed topology-aware embeddings (PDGNNs~\cite{Zhang2024Topology-awareembeddingmemoryforcl}). DeLoMe~\cite{DeLoMe} constructs a memory composed of lossless node prototypes, aiming to fully capture and store the graph information of old tasks, and further proposes a de-biased GCL loss function to address category imbalance. While these methods often demonstrate strong performance, their reliance on storing and re-processing historical data introduces significant computational and memory overhead, along with potential data privacy concerns.

\textbf{Prompt Learning and Parameter-Efficient Methods.}
An alternative direction is rooted in Parameter-Efficient Transfer Learning (PETL), a paradigm where large pre-trained models are adapted to new tasks by updating only a small subset of parameters. Representative methods like Adapters achieve this by inserting lightweight modules between the layers of a backbone network to reduce fine-tuning costs~\cite{houlsby2019parameterefficienttransferlearningnlp,Gao2024BeyondPL}.
Building on this foundation, prompt learning has emerged as a leading PETL technique. It guides a model's behavior for specific tasks by injecting learnable prompts,'' and is widely adopted for its excellent few-shot and transferability performance under thepre-train, prompt, predict'' framework~\cite{liu2021pretrainpromptpredictsystematic}. Inspired by its success, researchers have recently extended this paradigm to Graph Neural Networks (GNNs). For instance, UPT conditions a frozen GNN on structural prompts to enhance few-shot and cross-domain performance~\cite{Fang2022UniversalPT}, while GraphPrompt unifies pre-training and downstream tasks through a shared interface to improve generalization~\cite{Liu2023GraphPromptUP}. Other works like TPP infers task IDs to select the appropriate task-specific prompt for conditioning the GNN~\cite{niu2024replayandforgetfree}.
While these methods achieve impressive parameter-efficient knowledge transfer, their effectiveness critically depends on the availability of a powerful, often resource-intensive pre-trained backbone. This prerequisite poses a significant challenge in the graph domain, where universally effective and readily available pre-trained GNNs are not as common as in other fields like NLP or vision. This limitation motivates our research into a new paradigm that delivers the parameter efficiency of prompt-based conditioning, but without the dependency on a pre-trained model, thereby offering a more flexible and accessible solution for continual graph learning.
\section{Preliminary}

\begin{figure*}[t!]
  \centering
  \Description{A flowchart of the TAAM framework. It shows the training phase where a task-specific NSM and prototype are created, and the inference phase where the AMP method identifies the unknown task ID to select the correct expert module.}
  \includegraphics[width=0.80\linewidth]{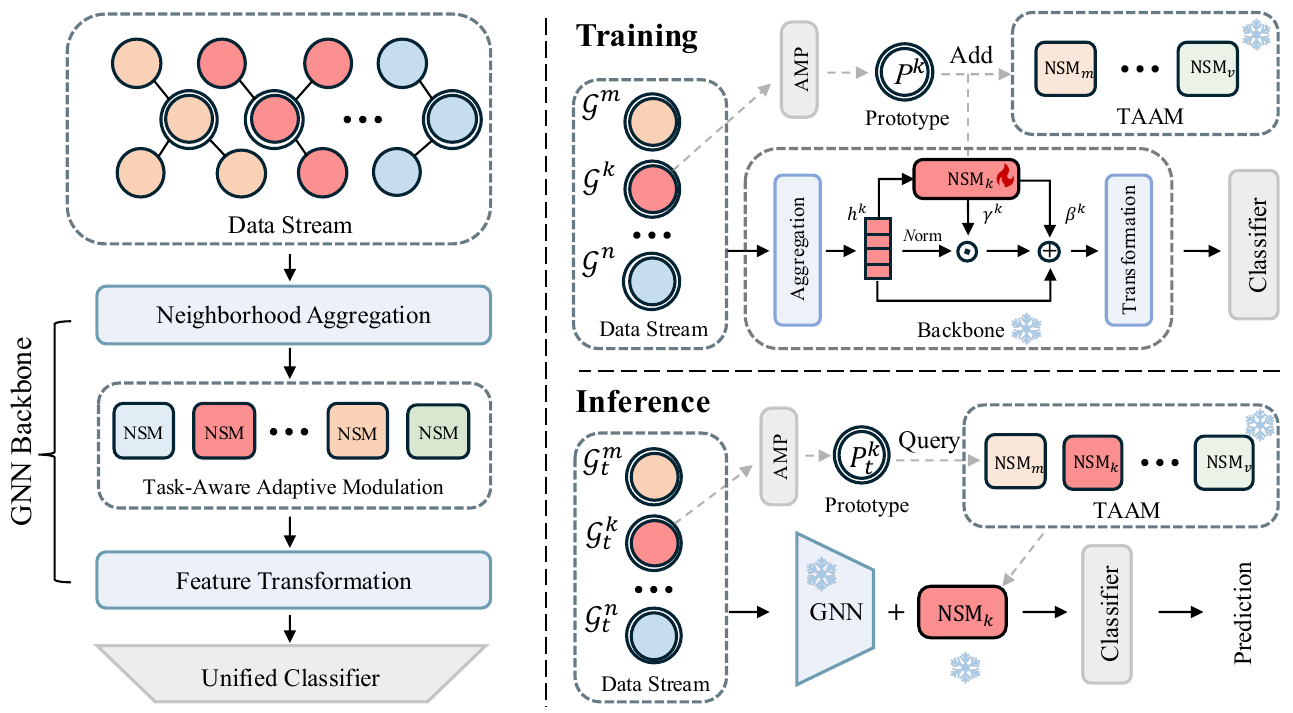}
  \caption{Overview of TAAM framework.TAAM is inserted the frozen GNN backbone,to steer a GNN's reasoning process in stream of graph data.The diagram illustrates the strategy of TAAM.Training Phase: For a new task $\mathcal{G}_k$, a new Neural Synapse Modulator ($\mathrm{NSM}_k$) is created from scratch and trained. Concurrently, its corresponding task prototype $P_k$ is generated and stored.After training is complete,  $\mathrm{NSM}_k$ frozen and added to the TAAM module bank.Inference Phase:For incoming test graph $\mathcal{G}^k_t$ with an unknown task ID , a prototype $P^k_t$ generated using AMP,for selects the most relevant ``expert" NSM.}
  \label{fig:overstructure}
\end{figure*} 

\subsection{Problem Formulation}

The work of this paper is oriented towards Graph Class-Incremental Learning (GCIL), where a model must continually learn new classes of nodes from a sequence of tasks. 

Formally, we are given a sequence of $N$ tasks, $\mathcal{T} = \{\tau_1, \tau_2, \dots, \tau_N\}$. Each task $\tau_k$ introduces a new set of node classes $\mathcal{C}_k$, where all class sets are disjoint, i.e., $\mathcal{C}_i \cap \mathcal{C}_j = \emptyset$ for any $i \neq j$. After learning up to task $\tau_k$, the model $f$ must be able to classify nodes from the union of all observed classes, $\mathcal{C}_{1:k} = \bigcup_{i=1}^k \mathcal{C}_i$. A critical constraint in the GCIL setting is that the task identity of data is \textbf{unavailable} during inference.

Unlike earlier GCL benchmarks~\cite{Zhang2022CGLBBT}, whichoperate in a transductive scenario (where training and testing nodes reside within the same graph), our study tackles the more demanding and practical \textbf{inductive} setting. Here, training and testing data for each task are fully isolated into independent subgraphs. Specifically, for each task $\tau_k$, the model is trained on a training set of graphs $\mathcal{G}{\text{train}}^k$ and evaluated on a distinct set of unseen test graphs $\mathcal{G}{\text{test}}^k$, where $\mathcal{G}{\text{train}}^k \cap \mathcal{G}{\text{test}}^k = \emptyset$. This setup challenges the model to generalize to new classes as well as to entirely novel graph structures and node instances that it has never encountered during training.

\subsection{Graph Neural Networks}

A GNN computes the hidden representation for a node $h_i^{(l)}$ at a given layer $l$ by aggregating features from its neighbors $\mathcal{N}(i)$\cite{gcn}. This process is generally defined by applying a transformation matrix $\mathbf{W}^{(l)}$ and a non-linear activation function $\sigma(\cdot)$:
\begin{equation}
    h_i^{(l)}=\sigma\left(\sum_{j\in\mathcal{N}(i)}\mathcal{A}_{ij}h_j^{(l-1)}\mathbf{W}^{(l)}\right)
\end{equation}
Here, $\mathcal{A}$ is a matrix defining the aggregation strategy, and $h_i^{(0)}$ is the initial node feature.
SGC ~\cite{Wu2019SimplifyingGC} streamlines the GNN architecture by removing intermediate non-linear activation functions. This decouples the model into two distinct stages:

\begin{enumerate}
    \item \textit{Neighborhood Aggregation} The initial node features $\mathbf{X}$ are aggregated over a K-hop neighborhood by being multiplied by the $K$-th power of the normalized adjacency matrix $\mathbf{S}$. This creates a fixed, pre-processed feature matrix $\mathbf{X}'$:
    \begin{equation}
        \mathbf{X}' = \mathbf{S}^K \mathbf{X}
    \end{equation}

    \item \textit{Feature Transformation:} A trainable Multi-Layer Perceptron (MLP) is then applied to the propagated features $\mathbf{X}'$.:
    \begin{equation}
        \hat{\mathbf{Y}} = \text{softmax}\left(\sigma\left(\mathbf{X}' \mathbf{W}_1\right) \mathbf{W}_2\right)
    \end{equation}
    where $\mathbf{W}_1$ and $\mathbf{W}_2$ are the trainable weight matrices of the MLP, and $\sigma(\cdot)$ is a non-linear activation function, such as ReLU.
\end{enumerate}

% %%
%
\section{Methodology}
\label{sec:methodology}

The proposed Task-Aware Adaptive Modulation (TAAM) framework offers a replay-free and resource-efficient solution for continual learning on graphs. TAAM utilizes a frozen Simple Graph Convolution (SGC) backbone as a stable feature extractor. For every new task in the sequence $\mathcal{T} = {\tau_1, \tau_2, \dots, \tau_N}$, it introduces a lightweight, task-specific Neural Synapse Modulator (NSM) to dynamically adapt the backbone’s output. Each task is assigned its own NSM (${\text{NSM}_1}, {\text{NSM}_2}, \dots, {\text{NSM}_N}$), as illustrated in Figure \ref{fig:overstructure}. At inference time, the system selects the appropriate NSM using robust task-ID inference, which is based on prototypes created by the Anchored Multi-hop Propagation (AMP) technique.

\subsection{Anchored Multi-hop Propagation for Task-ID Inference }
\label{sec:amp}

A critical component for stability in a continual learning setting without task labels at inference is the ability to accurately identify the task to which incoming data belongs. To achieve this, we propose generating a single, highly representative prototype vector $\mathbf{p}_k$ for each task $\tau_k$. 

We introduce \textbf{A}nchored \textbf{M}ulti-hop \textbf{P}ropagation (AMP), a method designed to create robust prototypes by capturing multi-scale neighborhood information. AMP leverages the principles of Approximate Personalized PageRank (APPNP)~\cite{Klicpera2018PredictTP}, which diffuses node features across the graph while retaining a connection to the initial features via a teleport probability $\alpha$. The propagation rule for the node feature matrix $\mathbf{H}$ is:
\begin{equation}
    \mathbf{Z}^{(i+1)} = (1-\alpha)\mathbf{S}\mathbf{Z}^{(i)} + \alpha\mathbf{H}^{(0)}
\end{equation}
where $\mathbf{Z}^{(0)} = \mathbf{H}^{(0)}$ are the initial node features, and $\mathbf{S}$ is the symmetrically normalized adjacency matrix.

Inspired by SIGN~\cite{Rossi2020SIGNSI}, we create a multi-scale representation by concatenating the diffused features from different propagation depths (hops). The final enriched feature matrix for prototype generation, $\mathbf{H}_{\text{AMP}}$, is formed as:
\begin{equation}
    \mathbf{H}_{\text{AMP}} = \text{CONCAT}(\mathbf{Z}^{(h_1)}, \mathbf{Z}^{(h_2)}, \dots, \mathbf{Z}^{(h_m)})
\end{equation}
where $\{h_1, \dots, h_m\}$ is the set of selected hops. The prototype $\mathbf{p}_k$ for task $\tau_k$ is then the mean of these enriched features over its training nodes $\mathcal{V}_k^{\text{train}}$:
\begin{equation}
    \mathbf{p}_k = \frac{1}{|\mathcal{V}_k^{\text{train}}|} \sum_{i \in \mathcal{V}_k^{\text{train}}} (\mathbf{H}_{\text{AMP}})_i
\end{equation}

During inference, we generate a prototype for the test data and identify the task ID by finding the most similar stored prototype using cosine similarity. The pseudocode for AMP is detailed in Algorithm~\ref{alg:amp}.

\begin{algorithm}[t]
\SetAlgoLined
\DontPrintSemicolon
\caption{Anchored Multi-hop Propagation}
\label{alg:amp}
\textbf{Input}: Graph $g$, Initial features $\mathbf{H}^{(0)}$, Training node set $\mathcal{V}_k^{\text{train}}$, AMP parameters ($L_{hops}, \alpha, K$);
\BlankLine
\textbf{Output}: Task prototype vector $\mathbf{p}_k$;
\BlankLine
\Begin{
    $\mathcal{H}_{collect} \gets \emptyset$;
    $\mathbf{Z} \gets \mathbf{H}^{(0)}$;
    \If{$0 \in L_{hops}$}{
        $\mathcal{H}_{collect} \gets \mathcal{H}_{collect} \cup \{\mathbf{Z}\}$;
    }
    \For{$i \gets 1$ \KwTo $K$}{
        $\mathbf{Z} \gets (1-\alpha)\mathbf{S}\mathbf{Z} + \alpha\mathbf{H}^{(0)}$; %\tcp*[r]{ref. Eq. (4)}
        \If{$i \in L_{hops}$}{
            $\mathcal{H}_{collect} \gets \mathcal{H}_{collect} \cup \{\mathbf{Z}\}$;
        }
    }
    $\mathbf{H}_{\text{AMP}} \gets \text{CONCAT}(\mathcal{H}_{collect})$; % \tcp*[r]{ref. Eq. (5)}    
    $\mathbf{p}_k \gets \frac{1}{|\mathcal{V}_k^{\text{train}}|} \sum_{v \in \mathcal{V}_k^{\text{train}}} (\mathbf{H}_{\text{AMP}})_v$; 
    \BlankLine
    \Return $\mathbf{p}_k$;
}
\end{algorithm}

%\tcp*[r]{ref. Eq. (6)}
\subsection{Neural Synapse Modulator}
\label{sec:nsm}

For each task $\tau_k$, we introduce a new, lightweight \textbf{N}eural \textbf{S}ynapse \textbf{M}odulator (NSM). The NSM acts as a task-specific expert that adapts the behavior of the frozen SGC backbone. It is a dynamic, node-attentive module that generates Feature-wise Linear Modulation (FiLM)~\cite{Brockschmidt2019GNNFiLMGN,Perez2017FiLMVR} parameters, $\boldsymbol{\gamma}$ and $\boldsymbol{\beta}$, for each node.

Specifically, the NSM for task $\tau_k$ contains a unique, learnable task embedding $\mathbf{e}_k \in \mathbb{R}^{D_{task}}$. To generate modulation parameters efficiently, we use a low-rank factorization. First, a set of $H$ base modulation vectors for $H$ attention heads, $\mathbf{M}_k \in \mathbb{R}^{H \times 2F'}$, is generated from the task embedding:
\begin{equation}
    \mathbf{M}_k = (\mathbf{W}_A \mathbf{e}_k) \cdot \mathbf{W}_B
\end{equation}
where $\mathbf{W}_A$ and $\mathbf{W}_B$ are learnable low-rank projection matrices.

Next, for each node $i$ with feature vector $\mathbf{h}_i$, we compute attention weights over the $H$ heads:
\begin{equation}
    \mathbf{a}_i = \text{softmax}(\mathbf{h}_i \mathbf{W}_{\text{att}})
\end{equation}
where $\mathbf{W}_{\text{att}} \in \mathbb{R}^{F \times H}$. The final per-node FiLM parameters are a weighted combination of the base modulations:
\begin{equation}
    [\boldsymbol{\gamma}_i, \boldsymbol{\beta}_i] = \mathbf{a}_i^\top \mathbf{M}_k
\end{equation}
These parameters are then used to modulate the node's feature vector $\mathbf{h}_i$ after the SGC's neighborhood aggregation step, followed by a residual connection:
\begin{equation}
    \mathbf{h}'_i = (\boldsymbol{\gamma}_i \odot \text{LayerNorm}(\mathbf{h}_i) + \boldsymbol{\beta}_i) + \mathbf{h}_i
\end{equation}
This node-attentive mechanism allows the NSM to provide fine-grained, context-specific adaptation of the backbone's representations.

\begin{figure*}[t!]
    \centering
    \Description{A heat map comparison showing performance matrices. TAAM maintains high accuracy (darker colors) along the diagonal and the lower triangle, demonstrating its ability to prevent catastrophic forgetting compared to baselines.}
    \includegraphics[width=1\linewidth]{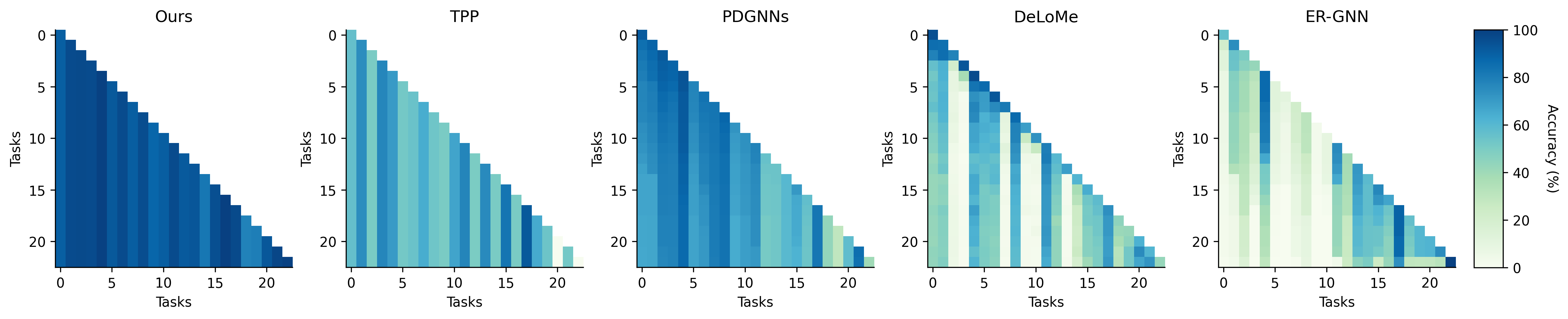}
      \caption{Performance matrices of differernt method on Products dataset.Darker colors indicate higher accuracy. TAAM (Ours) demonstrates consistently high accuracy across all tasks, indicating zero catastrophic forgetting.}
    \Description{}
  \label{fig:retention_matrix}
\end{figure*}

\subsection{Unified Classifier and Training Objective}

To accommodate the expanding set of classes, we employ a unified linear classifier that grows dynamically. Upon the arrival of a new task $\tau_k$ with class set $\mathcal{C}_k$, the classifier's weight matrix $\mathbf{W}_{k-1} \in \mathbb{R}^{D_{hid} \times C_{k-1}}$ is expanded to a new matrix $\mathbf{W}_k \in \mathbb{R}^{D_{hid} \times C_k}$, where $C_k = C_{k-1} + |\mathcal{C}_k|$. To preserve prior knowledge, the weights for existing classes are copied directly from $\mathbf{W}_{k-1}$, while only the weights corresponding to the new classes are randomly initialized. 

During the training phase for task $\tau_k$, we freeze all parameters except for those of the new NSM and the newly added weights in the classifier. This design strictly isolates parameter updates and prevents catastrophic forgetting at the decision layer.

To mitigate class imbalance, our training objective for task $\tau_k$ is a weighted cross-entropy loss. The weight for each class $c \in \mathcal{C}_k$ is its inverse frequency in the training data, $w_c = 1/N_c$, where $N_c$ is the number of training instances for class $c$. Crucially, the loss is computed exclusively over the set of new classes $\mathcal{C}_k$:
\begin{equation}
\label{eq:training_loss}
\mathcal{L}_k = - \sum_{v \in \mathcal{V}_k^{\text{train}}} w_{y_v} \log\left(\frac{\exp(\mathbf{z}_{v, y_v})}{\sum_{c' \in \mathcal{C}_k} \exp(\mathbf{z}_{v, c'})}\right)
\end{equation}
where $\mathcal{V}_k^{\text{train}}$ are the training nodes for task $\tau_k$, $\mathbf{z}_v$ is the full logit vector for node $v$, and $y_v \in \mathcal{C}_k$ is its corresponding label.
\begin{table*}[t!]
  \centering
  \caption{Comparison of different models on different data sets in the GCIL setting. The bold results are the best performance excluding Joint and Oracle. A `$\uparrow$` indicates that a greater value represents better performance, while a `$\downarrow$` indicates the opposite. `Joint` can access the data of all tasks; `Oracle` can access the data and task IDs of all tasks.}
  \label{tab:main_results}
  \scriptsize % Use scriptsize font for the entire table
  \setlength{\tabcolsep}{2.5pt} % Further reduce space between columns for the smaller font
  \begin{tabular}{llcccccccccccccccc}
    \toprule
    \multirow{2}{*}{\textbf{Category}} & \multirow{2}{*}{\textbf{Model}} & \multicolumn{2}{c}{\textbf{CoraFull}} & \multicolumn{2}{c}{\textbf{Arxiv}} & \multicolumn{2}{c}{\textbf{Reddit}} & \multicolumn{2}{c}{\textbf{Products}} & \multicolumn{2}{c}{\textbf{Photo}} & \multicolumn{2}{c}{\textbf{WikiCS}} & \multicolumn{2}{c}{\textbf{Computer}} & \multicolumn{2}{c}{\textbf{Citeseer}} \\
    \cmidrule(lr){3-4} \cmidrule(lr){5-6} \cmidrule(lr){7-8} \cmidrule(lr){9-10} \cmidrule(lr){11-12} \cmidrule(lr){13-14} \cmidrule(lr){15-16} \cmidrule(lr){17-18}
    & & AA\%($\uparrow$) & AF\%($\downarrow$) & AA\%($\uparrow$) & AF\%($\downarrow$) & AA\%($\uparrow$) & AF\%($\downarrow$) & AA\%($\uparrow$) & AF\%($\downarrow$) & AA\%($\uparrow$) & AF\%($\downarrow$) & AA\%($\uparrow$) & AF\%($\downarrow$) & AA\%($\uparrow$) & AF\%($\downarrow$) & AA\%($\uparrow$) & AF\%($\downarrow$) \\
    \midrule
    \multirow{2}{*}{Lower bound} & Finetuning & 1.8\pms{0.2} & 13.7\pms{2.5} & 4.9\pms{0.0} & 11.8\pms{4.1} & 4.9\pms{0.0} & 4.9\pms{0.0} & 4.7\pms{0.8} & 31.1\pms{1.6} & 23.2\pms{0.0} & 38.6\pms{6.5} & 18.8\pms{0.1} & 41.5\pms{7.9} & 19.0\pms{0.1} & 28.1\pms{12.0} & 31.4\pms{0.1} & 52.3\pms{5.3} \\
    & LwF(2017) & 1.8\pms{0.2} & 18.2\pms{2.1} & 4.9\pms{0.0} & 30.8\pms{4.8} & 5.6\pms{0.9} & 58.4\pms{5.3} & 21.4\pms{0.9} & 33.4\pms{0.7} & 12.5\pms{8.0} & 46.2\pms{3.1} & 14.7\pms{1.2} & 78.4\pms{6.0} & 8.5\pms{7.8} & 41.8\pms{32.0} & 31.4\pms{0.1} & 70.7\pms{0.8} \\
    \midrule
    \multirow{3}{*}{Regularisation} & MAS(2018) & 2.8\pms{0.3} & 67.4\pms{1.4} & 4.8\pms{0.0} & 80.9\pms{4.2} & 5.0\pms{0.1} & 97.5\pms{0.1} & 3.9\pms{0.3} & 88.7\pms{0.4} & 18.6\pms{0.1} & 92.3\pms{2.0} & 18.6\pms{0.1} & 92.3\pms{1.8} & 23.1\pms{0.2} & 87.3\pms{14.2} & 29.9\pms{0.1} & 80.9\pms{0.4} \\
    & TWP(2021) & 7.2\pms{0.9} & 65.7\pms{1.3} & 4.9\pms{0.0} & 60.5\pms{5.3} & 5.0\pms{0.3} & 92.8\pms{1.5} & 5.5\pms{0.2} & 78.3\pms{1.7} & 23.2\pms{0.1} & 73.8\pms{7.5} & 18.6\pms{0.1} & 89.2\pms{2.5} & 19.2\pms{0.3} & 64.7\pms{15.4} & 36.5\pms{0.4} & 66.9\pms{1.2} \\
    & ER-GNN(2021) & 2.1\pms{0.0} & 14.1\pms{1.8} & 16.3\pms{2.7} & 40.4\pms{2.4} & 85.0\pms{0.8} & 12.3\pms{0.8} & 22.5\pms{0.9} & 33.1\pms{1.2} & 94.3\pms{0.2} & 2.4\pms{0.2} & 84.1\pms{0.3} & 7.3\pms{0.5} & 87.6\pms{0.6} & 10.6\pms{0.9} & 61.3\pms{0.4} & 34.5\pms{0.6} \\
    \midrule
    \multirow{3}{*}{Replay} & SSM(2022) & 60.3\pms{0.6} & 11.2\pms{0.3} & 7.7\pms{2.1} & 8.8\pms{2.5} & 91.9\pms{0.3} & 1.5\pms{0.2} & 32.9\pms{1.5} & 15.3\pms{1.6} & 92.6\pms{0.7} & 1.9\pms{0.5} & 74.3\pms{1.6} & 4.4\pms{1.1} & 88.7\pms{0.9} & 3.1\pms{0.8} & 47.1\pms{4.5} & 6.5\pms{2.2} \\
    & PDGNNs(2024) & 49.8\pms{27.0} & 11.5\pms{7.9} & 44.5\pms{0.2} & 14.2\pms{0.1} & 94.9\pms{0.1} & 1.4\pms{0.0} & 45.2\pms{0.1} & 7.1\pms{0.4} & 94.6\pms{0.3} & 2.2\pms{0.2} & 84.2\pms{0.5} & 3.8\pms{0.4} & 91.4\pms{0.8} & 5.5\pms{0.9} & 71.0\pms{0.1} & 10.9\pms{0.1} \\
    & DeLoMe(2024) & 30.5\pms{0.1} & 47.0\pms{0.5} & 31.0\pms{0.2} & 24.1\pms{0.2} & 94.1\pms{0.6} & 0.2\pms{0.1} & 37.8\pms{0.5} & 35.8\pms{0.2} & 47.4\pms{0.2} & 64.9\pms{0.3} & 35.6\pms{0.3} & 67.1\pms{0.3} & 37.1\pms{0.1} & 73.7\pms{0.2} & 48.4\pms{0.2} & 32.3\pms{0.2} \\
    \midrule
    \multirow{2}{*}{\shortstack{Parameter\\isolation}} & TPP(2024) & 88.9\pms{0.8} & 0.0\pms{0.0} & 84.4\pms{1.0} & 0.0\pms{0.0} & 98.5\pms{0.2} & 0.0\pms{0.0} & 84.9\pms{0.2} & 0.0\pms{0.0} & 23.3\pms{0.7} & 0.0\pms{0.0} & 36.6\pms{0.5} & 0.0\pms{0.0} & 19.9\pms{0.1} & 46.2\pms{0.5} & 77.7\pms{1.6} & 0.0\pms{0.0} \\
    & \textbf{TAAM (Ours)} & \textbf{92.5\pms{0.0}} & \textbf{0.0\pms{0.0}} & \textbf{88.6\pms{0.4}} & \textbf{0.0\pms{0.0}} & \textbf{99.0\pms{0.0}} & \textbf{0.0\pms{0.0}} & \textbf{92.9\pms{0.5}} & \textbf{0.0\pms{0.0}} & \textbf{96.2\pms{0.4}} & \textbf{0.0\pms{0.0}} & \textbf{92.8\pms{0.3}} & \textbf{0.0\pms{0.0}} & \textbf{97.5\pms{0.6}} & \textbf{0.0\pms{0.0}} & \textbf{83.5\pms{1.0}} & \textbf{0.0\pms{0.0}} \\
    \midrule
    \multirow{2}{*}{Upper bound} & Joint & 64.1\pms{0.3} & - & 45.2\pms{0.1} & - & 95.6\pms{0.1} & - & 65.2\pms{0.3} & - & 95.1\pms{0.5} & - & 83.7\pms{0.3} & - & 94.9\pms{0.1} & - & 70.7\pms{0.1} & - \\
    & Oracle & 94.3\pms{0.3} & - & 88.5\pms{0.3} & - & 99.2\pms{0.0} & - & 94.0\pms{0.6} & - & 97.0\pms{0.1} & - & 92.9\pms{0.2} & - & 97.7\pms{0.1} & - & 85.2\pms{0.2} & - \\
    \bottomrule
  \end{tabular}
\end{table*}

\section{Experiments}
In this section, we conduct a comprehensive empirical evaluation to validate the effectiveness and efficiency of our proposed Task-Aware Adaptive Modulation (TAAM) framework. Our experiments are designed to answer four key research questions: 
\begin{enumerate}
    \item[\textbf{Q1:}] How does TAAM perform against state-of-the-art (SOTA) Graph Continual Learning (GCL) methods in terms of accuracy and catastrophic forgetting?
    \item[\textbf{Q2:}] How efficient is TAAM concerning computational time, memory usage, and parameter overhead?
    \item[\textbf{Q3:}] How crucial is TAAM's task-ID inference mechanism, and how does it compare to existing approaches?
    \item[\textbf{Q4:}] What is the specific contribution of each core component within the TAAM architecture?
\end{enumerate}

\subsection{Experimental Setup}

\paragraph{Datasets and Baselines.}
We perform evaluations on eight benchmark datasets, which span several domains. Specifically, these include three citation networks: CoraFull~\cite{McCallum2000AutomatingTC}, Arxiv~\cite{Hu2020OpenGB}, and Citeseer~\cite{Yang2016RevisitingSL}; one social network: Reddit~\cite{Hamilton2017InductiveRL}; three product co-purchase networks: Products~\cite{Hu2020OpenGB}, Photo~\cite{Aozm}, and Computer~\cite{Aozm}; and one webpage-based network: WikiCS~\cite{liu2023one}, with statistics detailed in Table~\ref{tab:dataset_stats}. Our baselines cover the spectrum of GCL strategies,including regularization, replay-base, and parameter isolation leading methods. We use Joint training and Oracle models as upper bounds.

\paragraph{Implementation Details.}

To ensure a rigorous and practical evaluation, our experimental setup adapts the CGL benchmark~\cite{Zhang2022CGLBBT} to a strict \textbf{inductive learning scenario}. In this setting, the training, validation, and test sets are constructed from entirely disjoint subgraphs. This deliberate separation is a crucial correction to prior evaluation protocols, as it prevents any potential data leakage from the test distribution during the training phase. Consequently, this setup guarantees a more realistic and unbiased assessment of a model's generalization capabilities in a continual learning environment.

For a fair comparison across all methods, we employ a unified backbone feature extractor: a two-layer Simplifying Graph Convolutional Network (SGC)~\cite{Wu2019SimplifyingGC} with a 256-dimensional hidden layer. The SGC is configured without batch normalization, dropout, or bias. Its parameters are randomly initialized and subsequently frozen throughout the entire continual learning process, ensuring that all performance gains come from the continual learning modules themselves.

All models were trained for 200 epochs per task using the Adam optimizer, with a learning rate of $0.005$ and a weight decay of $5 \times 10^{-4}$. We use the same set of random seeds for all experiments to ensure reproducibility. Hyperparameters for all baseline methods were configured according to their original publications. For our proposed model, TAAM, the dimension of the learnable task embedding, $\mathbf{e}_{\tau}$, is set to 6. The internal node-attentive mechanism in each Node-Specific Modulator (NSM) uses $K=2$ heads. Critically, for each new task, only the parameters of the newly instantiated NSM and its corresponding classifier are trainable, embodying our lightweight learning paradigm.

\paragraph{Evaluation Metrics.}
We evaluate performance using Average Accuracy (AA) and Average Forgetting (AF), where a higher AA indicates better overall accuracy and a lower AF indicates less catastrophic forgetting.

\begin{table}[t]
  \centering
  \caption{Statistics of the datasets.}
  \label{tab:dataset_stats}
  \small % Use a smaller font size for the entire table
  \setlength{\tabcolsep}{4pt} % Reduce the space between columns
  \begin{tabular}{lrrrrr}
    \toprule
    \textbf{Dataset} & \textbf{Nodes} & \textbf{Edges} & \textbf{Feats.} & \textbf{Classes} & \textbf{Tasks} \\
    \midrule
    Citeseer   & 3,327       & 9,228         & 3,703 & 6  & 3  \\
    Photo      & 48,362      & 500,928       & 745   & 8  & 4  \\
    WikiCS     & 11,701      & 431,726       & 300   & 10 & 5  \\
    Computer   & 13,752      & 491,722       & 767   & 10 & 5  \\
    CoraFull   & 19,793      & 130,622       & 8,710 & 70 & 35 \\
    Arxiv      & 169,343     & 1,166,243     & 128   & 40 & 20 \\
    Reddit     & 232,965     & 114,615,892   & 602   & 40 & 20 \\
    Products   & 2,449,029   & 61,859,036    & 100   & 47 & 23 \\
    \bottomrule
  \end{tabular}
\end{table}

\subsection{Main results}

\paragraph{State-of-the-Art Performance (\textbf{Q1})}
As shown in Table~\ref{tab:main_results}, TAAM establishes a new state-of-the-art by achieving the highest Average Accuracy (AA) across all eight datasets while simultaneously eliminating catastrophic forgetting (0.0\% AF). This superiority stems from its unique architectural design, which overcomes the limitations of prior paradigms. Unlike replay-based methods, TAAM avoids their inherent computational and privacy overhead while still delivering dominant performance on large-scale graphs like Products (92.9\% vs. 45.2\% for PDGNNs). More critically, its advancement over other parameter-isolation methods like TPP is driven by its novel Task-Aware module and fine-grained Neural Synapse Modulators (NSMs). This mechanism provides superior adaptability, leading to remarkable performance gains such as the 77.6\% improvement on the Computer dataset. Ultimately, TAAM presents a more effective solution to the \textbf{stability-plasticity trade-off}: its core framework guarantees perfect stability (zero forgetting), while its adaptive modulators deliver SOTA plasticity.

\paragraph{Visualizing Knowledge Retention}
To provide a more intuitive understanding of this zero-forgetting property, we visualize the model's task-wise performance after the final training phase. As depicted in Figure~\ref{fig:retention_matrix}, the heatmaps vividly illustrate our method's superior knowledge retention. The matrix for TAAM (Ours) exhibits a consistently dark and uniform color across the lower triangle. This signifies both excellent plasticity (the dark diagonal, indicating high accuracy on newly learned tasks) and perfect stability (no color degradation in the off-diagonal elements, confirming zero forgetting). In stark contrast, the heatmaps for replay-based methods like DeLoMe and regularization-based ER-GNN reveal a clear pattern of knowledge decay. While their diagonals indicate an ability to learn, the pronounced color fading down each column provides clear visual evidence of catastrophic forgetting, corroborating the high AF\% values reported in Table~\ref{tab:main_results}.

\subsection{Efficiency Analysis}

\paragraph{Efficiency Analysis (\textbf{Q2})}
A core design philosophy of TAAM is resource efficiency, a critical factor for practical GCL deployment. We analyze this through the lens of the performance-cost trade-off, considering parameters, memory, and time.

\begin{figure}[t]
    \centering
    \Description{A scatter plot or line chart comparing the training time efficiency and Average Accuracy (AA) between TAAM and other state-of-the-art graph continual learning methods.}
    \includegraphics[width=1\columnwidth]{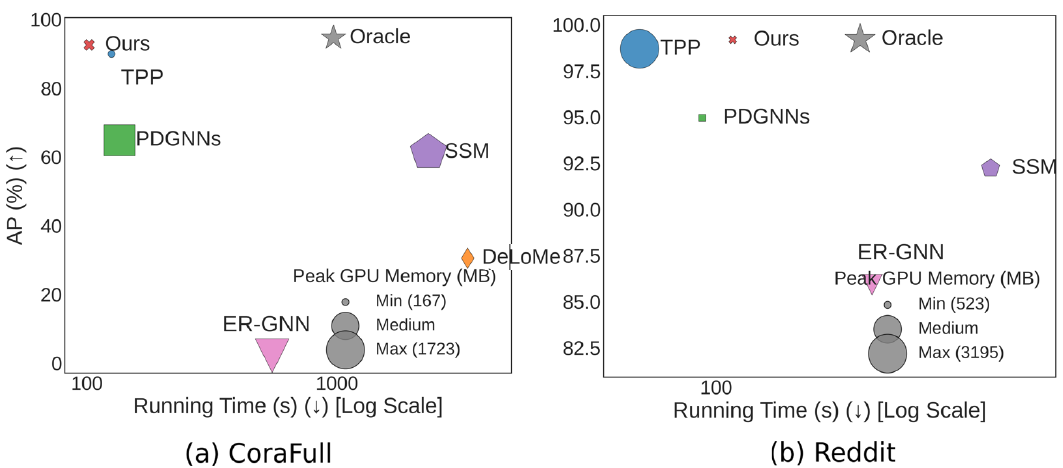}
    \caption{Comparison of running time and AA}
    \label{fig:tradeoff2}
\end{figure}

As illustrated in Figure~\ref{fig:tradeoff2}, the optimal efficiency profile—high accuracy with low computational cost—resides in the top-left quadrant. \textbf{TAAM is consistently positioned in this optimal region}, achieving state-of-the-art accuracy with minimal running time and memory footprint. However, this plot only reveals part of the story. For instance, while TPP appears competitive in inference time, this view conceals the \textbf{heavy computational overhead of its mandatory pre-training phase}. Similarly, some methods are omitted entirely for clarity due to their extreme inefficiency; \textbf{DeLoMe}, for example, requires substantial time to generate node prototypes, with a single experiment on the large-scale Reddit dataset consuming approximately one hour.

\paragraph{Parameter Efficiency}
This computational efficiency is rooted in our model's design. As shown in Table~\ref{tab:param_efficiency}, TAAM demonstrates exceptional parameter efficiency by introducing only 0.02M - 3.08M additional parameters per task (measured by the number of float32 values), which is orders of magnitude lower than methods like ER-GNN and PDGNNs. This makes TAAM a truly lightweight solution. For scenarios where peak performance is paramount, our variant \textbf{TAAM-L} offers a flexible trade-off, further boosting accuracy with only a modest increase in parameters (Table~\ref{tab:ablation_results}), demonstrating the adaptability of our framework to different efficiency-performance requirements.

\begin{figure}[t]
    \centering
    \Description{A bar chart comparing the task ID prediction accuracy of our proposed Anchored Multi-hop Propagation (AMP) method against the Laplacian smoothing (LS) baseline.}
    \includegraphics[width=1\columnwidth]{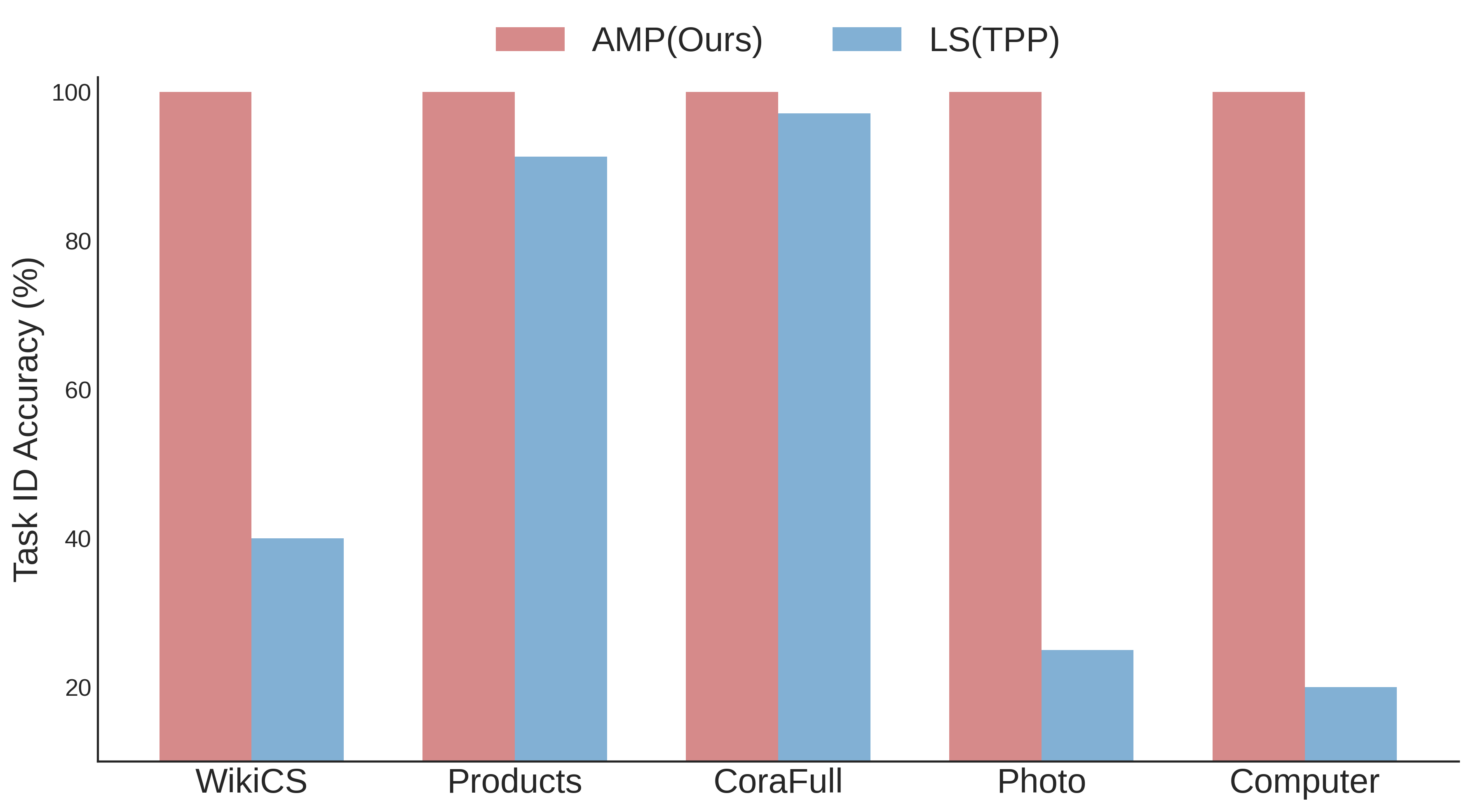}
    \caption{Task ID prediction accuracy of AMP and Laplacian smoothing (LS) of TPP}
    \label{fig:task_id_acc}
\end{figure}

\begin{table*}[t!]
\centering
\caption{Ablation results of TAAM and its variants.}

\label{tab:ablation_results}
\scriptsize
\setlength{\tabcolsep}{3pt}
\begin{tabular}{lcccccccccccccccc}
\toprule
\multirow{2}{*}{\textbf{Ablated Components}} & \multicolumn{2}{c}{\textbf{CoraFull}} & \multicolumn{2}{c}{\textbf{Arxiv}} & \multicolumn{2}{c}{\textbf{Reddit}} & \multicolumn{2}{c}{\textbf{Products}} & \multicolumn{2}{c}{\textbf{Photo}} & \multicolumn{2}{c}{\textbf{WikiCS}} & \multicolumn{2}{c}{\textbf{Computer}} & \multicolumn{2}{c}{\textbf{Citeseer}} \\
\cmidrule(lr){2-3} \cmidrule(lr){4-5} \cmidrule(lr){6-7} \cmidrule(lr){8-9} \cmidrule(lr){10-11} \cmidrule(lr){12-13} \cmidrule(lr){14-15} \cmidrule(lr){16-17}
& AA\%($\uparrow$) & AF\%($\downarrow$) & AA\%($\uparrow$) & AF\%($\downarrow$) & AA\%($\uparrow$) & AF\%($\downarrow$) & AA\%($\uparrow$) & AF\%($\downarrow$) & AA\%($\uparrow$) & AF\%($\downarrow$) & AA\%($\uparrow$) & AF\%($\downarrow$) & AA\%($\uparrow$) & AF\%($\downarrow$) & AA\%($\uparrow$) & AF\%($\downarrow$) \\
\midrule
w/o Task-Aware & 2.0$\pm$0.3 & 5.0$\pm$0.9 & 4.9$\pm$0.0 & 14.8$\pm$2.4 & 5.3$\pm$0.9 & 27.9$\pm$4.6 & 2.7$\pm$0.5 & 11.6$\pm$3.0 & 21.2$\pm$0.8 & 55.6$\pm$10.5 & 18.9$\pm$0.1 & 32.1$\pm$14.2 & 18.9$\pm$0.1 & 32.3$\pm$14.6 & 29.7$\pm$0.6 & 37.4$\pm$14.4 \\
w/o NSMs & 5.8$\pm$0.8 & 0.0$\pm$0.0 & 7.8$\pm$2.8 & 0.0$\pm$0.0 & 9.0$\pm$3.7 & 0.0$\pm$0.0 & 8.8$\pm$2.0 & 0.0$\pm$0.0 & 30.6$\pm$7.2 & 0.0$\pm$0.0 & 23.5$\pm$9.1 & 0.0$\pm$0.0 & 25.5$\pm$9.0 & 0.0$\pm$0.0 & 31.7$\pm$4.5 & 0.0$\pm$0.0 \\
w/o Unified Classifier & 92.2$\pm$0.5 & 0.0$\pm$0.0 & 88.3$\pm$0.4 & 0.0$\pm$0.0 & 99.0$\pm$0.0 & 0.0$\pm$0.0 & 93.0$\pm$0.5 & 0.0$\pm$0.0 & 95.7$\pm$0.7 & 0.0$\pm$0.0 & 92.9$\pm$0.2 & 0.0$\pm$0.0 & 97.7$\pm$0.5 & 0.0$\pm$0.0 & 82.0$\pm$0.8 & 0.0$\pm$0.0 \\
\midrule
TAAM & 92.5$\pm$0.5 & 0.0$\pm$0.0 & 88.6$\pm$0.4 & 0.0$\pm$0.0 & 99.0$\pm$0.0 & 0.0$\pm$0.0 & 92.9$\pm$0.5 & 0.0$\pm$0.0 & 96.2$\pm$0.4 & 0.0$\pm$0.0 & 92.8$\pm$0.3 & 0.0$\pm$0.0 & 97.5$\pm$0.6 & 0.0$\pm$0.0 & 83.5$\pm$1.0 & 0.0$\pm$0.0 \\
TAAM-L (w/o LoRA) & 93.6$\pm$0.3 & 0.0$\pm$0.0 & 87.9$\pm$0.5 & 0.0$\pm$0.0 & 99.1$\pm$0.1 & 0.0$\pm$0.0 & 93.0$\pm$0.6 & 0.0$\pm$0.0 & 95.9$\pm$0.7 & 0.0$\pm$0.0 & 93.4$\pm$0.2 & 0.0$\pm$0.0 & 98.0$\pm$0.4 & 0.0$\pm$0.0 & 84.9$\pm$0.3 & 0.0$\pm$0.0 \\
\bottomrule
\end{tabular}
\end{table*}

\begin{table}[t]
  \centering
  \caption{Parameter efficiency comparison.}
  \label{tab:param_efficiency}
  \begin{tabular}{lcrr}
    \toprule
    % "Method" and "Total Avg AA" now span 2 rows
    \multirow{2}{*}{\textbf{Method}} & \multirow{2}{*}{\textbf{Total Avg AA($\uparrow$)}} & \multicolumn{2}{c}{\textbf{Additional Cost}} \\
    \cmidrule(lr){3-4}
                                     &                                                   & Min.      & Max.\\
    \midrule
    Oracle                           & 93.6                                              & 2.12 M             & 82.5638 M \\
    ER-GNN                           & 56.7                                              & 1.80 M             & 103.2285 M \\
    SSM                              & 61.9                                              & 0.30 M             & 51.6099 M \\
    DeLoMe                           & 45.2                                              & 0.18 M             & 34.4037 M \\
    PDGNNs                           & 71.9                                              & 1.06 M             & 100.2838 M \\
    TPP                              & 64.3                                              & 0.01 M             & 1.7949 M \\
    \midrule
    TAAM-L                           & 93.2                                              & 0.05 M             & 9.1812 M \\
    TAAM                    & 92.9                                     & 0.02 M   & 3.0861 M \\
    \bottomrule
  \end{tabular}
\end{table}

\subsection{Analysis of Core Components}

\paragraph{Superior Task-ID Inference (\textbf{Q3})}
Reliable task-ID inference is the foundation of our replay-free approach. We compare our Anchored Multi-hop Propagation (AMP) against the Laplacian smoothing (LS) method used in TPP. Figure~\ref{fig:task_id_acc} shows that AMP achieves near-perfect task-ID accuracy across diverse datasets, significantly outperforming LS. This robust inference capability prevents error propagation and is key to TAAM's stability and consistent performance.

\paragraph{Ablation Study (\textbf{Q4})}

To dissect the model architecture, we conducted a comprehensive ablation study, The key findings from our ablation study are as follows:

\begin{itemize}
    \item \textbf{w/o Task-Aware Module}: The removal of this component leads to a catastrophic collapse in performance. This confirms that the task-aware module is the cornerstone of our architecture, enabling the frozen GNN backbone to effectively adapt to diverse task distributions.
    
    \item \textbf{w/o NSMs}: Excluding the NSMs results in a severe degradation in accuracy, even though forgetting is not induced. This validates our hypothesis that fine-grained, instance-level adaptation is crucial for capturing the unique characteristics of nodes within each task.
    
    \item \textbf{w/o Unified Classifiers}: Compared with using independent classifiers for each task, adopting a unified classifier will lead to a persistent slight decline in performance, which indicates that decoupled classification heads can effectively avoid interference between tasks..
    
    \item \textbf{TAAM vs. TAAM-L (w/o LoRA)}:  A core design choice in our framework is the integration of LoRA to parameterize the NSMs. This is a deliberate  trade-off:  its primary purpose is to ensure that the number of new parameters per task remains minimal,  preventing the model size from growing uncontrollably as more tasks are learned. To validate this strategy,  we compare our full TAAM model against TAAM-L,  a variant employing a simpler modulator without the LoRA parameterization. The results show that complete model is slightly better than that of the variant without LoRA It has been proven that LoRA has achieved an excellent balance between ensuring parameter efficiency and maintaining model performance.
\end{itemize}

\subsection{Parameter Efficiency}
Finally, we analyze the parameter efficiency of TAAM. Table~\ref{tab:param_efficiency} compares the additional parameter cost required by different continual learning methods. It is evident that TAAM and its variant TAAM-L are exceptionally parameter-efficient. Our method requires only 0.02M to 3.08M additional parameters, a cost that is orders of magnitude lower than most replay-based and regularization-based methods like ER-GNN and PDGNNs, and even more efficient than the parameter-isolation baseline TPP. This demonstrates that TAAM not only achieves superior performance but does so with remarkable efficiency, making it a practical and scalable solution for real-world graph continual learning scenarios.

\section{Conclusion}
\label{conclusion}

In this paper, we introduce TAAM, a new replay-free framework for Graph Continual Learning (GCL) designed to address the stability-plasticity dilemma without the memory and privacy issues associated with replay-based methods. The core of TAAM balances performance and parameter efficiency: it combines a frozen GNN backbone, which maintains prior knowledge, with lightweight, task-specific Neural Synapse Modulators (NSMs) that are trained and kept fixed for each new task. This design intentionally adds only a small number of parameters per task, preventing the model size from growing uncontrollably while still allowing enough plasticity to learn new concepts. Additionally, we propose Anchored Multi-hop Propagation (AMP), a theoretically grounded method to infer task IDs in real-world scenarios where they are unknown. Importantly, we identify flaws in existing GCL benchmarks and perform a thorough evaluation in a more rigorous and realistic inductive learning setting. Extensive experiments across eight datasets show that TAAM consistently and significantly outperforms state-of-the-art methods.

%%%%%%%%%%%%%%%%%%%%%%%%%%%%%%%%%%%%%%%%%%%%%%%%%%%%%%%%%%%%%%%%%%%%%%%%

%%% The next two lines define, first, the bibliography style to be 
%%% applied, and, second, the bibliography file to be used.
% \clearpage
\bibliographystyle{ACM-Reference-Format}
\balance 
\bibliography{sample}

@inproceedings{Su2024TowardsRG,
  author       = {Junwei Su and
                  Difan Zou and
                  Zijun Zhang and
                  Chuan Wu},
  editor       = {Andreas Krause and
                  Emma Brunskill and
                  Kyunghyun Cho and
                  Barbara Engelhardt and
                  Sivan Sabato and
                  Jonathan Scarlett},
  title        = {Towards Robust Graph Incremental Learning on Evolving Graphs},
  booktitle    = {International Conference on Machine Learning, {ICML} 2023, 23-29 July
                  2023, Honolulu, Hawaii, {USA}},
  series       = {Proceedings of Machine Learning Research},
  volume       = {202},
  pages        = {32728--32748},
  publisher    = {{PMLR}},
  year         = {2023},
  url          = {https://proceedings.mlr.press/v202/su23a.html},
  bibsource    = {dblp computer science bibliography, https://dblp.org}
}

@inproceedings{Tian2024ContinualLO,
  author       = {Zonggui Tian and
                  Du Zhang and
                  Hong{-}Ning Dai},
  title        = {Continual Learning on Graphs: {A} Survey},
  journal      = {CoRR},
  volume       = {abs/2402.06330},
  year         = {2024},
  url          = {https://doi.org/10.48550/arXiv.2402.06330},
  doi          = {10.48550/ARXIV.2402.06330},
  eprinttype    = {arXiv},
  eprint       = {2402.06330},
  bibsource    = {dblp computer science bibliography, https://dblp.org}
}

@inproceedings{Yang2016RevisitingSL,
  author       = {Zhilin Yang and
                  William W. Cohen and
                  Ruslan Salakhutdinov},
  editor       = {Maria{-}Florina Balcan and
                  Kilian Q. Weinberger},
  title        = {Revisiting Semi-Supervised Learning with Graph Embeddings},
  booktitle    = {Proceedings of the 33nd International Conference on Machine Learning,
                  {ICML} 2016, New York City, NY, USA, June 19-24, 2016},
  series       = {{JMLR} Workshop and Conference Proceedings},
  volume       = {48},
  pages        = {40--48},
  publisher    = {JMLR.org},
  year         = {2016},
  url          = {http://proceedings.mlr.press/v48/yanga16.html},
  bibsource    = {dblp computer science bibliography, https://dblp.org}
}

@inproceedings{Hamilton2017InductiveRL,
  author       = {William L. Hamilton and
                  Zhitao Ying and
                  Jure Leskovec},
  editor       = {Isabelle Guyon and
                  Ulrike von Luxburg and
                  Samy Bengio and
                  Hanna M. Wallach and
                  Rob Fergus and
                  S. V. N. Vishwanathan and
                  Roman Garnett},
  title        = {Inductive Representation Learning on Large Graphs},
  booktitle    = {Advances in Neural Information Processing Systems 30: Annual Conference
                  on Neural Information Processing Systems 2017, December 4-9, 2017,
                  Long Beach, CA, {USA}},
  pages        = {1024--1034},
  year         = {2017},
  url          = {https://proceedings.neurips.cc/paper/2017/hash/5dd9db5e033da9c6fb5ba83c7a7ebea9-Abstract.html},
  bibsource    = {dblp computer science bibliography, https://dblp.org}
}

@inproceedings{McCallum2000AutomatingTC,
  author       = {Andrew Kachites McCallum and
                  Kamal Nigam and
                  Jason Rennie and
                  Kristie Seymore},
  title        = {Automating the Construction of Internet Portals with Machine Learning},
  journal      = {Inf. Retr.},
  volume       = {3},
  number       = {2},
  pages        = {127--163},
  year         = {2000},
  url          = {https://doi.org/10.1023/A:1009953814988},
  doi          = {10.1023/A:1009953814988},
  bibsource    = {dblp computer science bibliography, https://dblp.org}
}

@inproceedings{Zhang2022CGLBBT,
  author       = {Xikun Zhang and
                  Dongjin Song and
                  Dacheng Tao},
  editor       = {Sanmi Koyejo and
                  S. Mohamed and
                  A. Agarwal and
                  Danielle Belgrave and
                  K. Cho and
                  A. Oh},
  title        = {{CGLB:} Benchmark Tasks for Continual Graph Learning},
  booktitle    = {Advances in Neural Information Processing Systems 35: Annual Conference
                  on Neural Information Processing Systems 2022, NeurIPS 2022, New Orleans,
                  LA, USA, November 28 - December 9, 2022},
  year         = {2022},
  url          = {http://papers.nips.cc/paper\_files/paper/2022/hash/548a41b9cac6f50dccf7e63e9e1b1b9b-Abstract-Datasets\_and\_Benchmarks.html},
  bibsource    = {dblp computer science bibliography, https://dblp.org}
}

@inproceedings{Wang2020LifelongGL,
  author       = {Chen Wang and
                  Yuheng Qiu and
                  Dasong Gao and
                  Sebastian A. Scherer},
  title        = {Lifelong Graph Learning},
  booktitle    = {{IEEE/CVF} Conference on Computer Vision and Pattern Recognition,
                  {CVPR} 2022, New Orleans, LA, USA, June 18-24, 2022},
  pages        = {13709--13718},
  publisher    = {{IEEE}},
  year         = {2022},
  url          = {https://doi.org/10.1109/CVPR52688.2022.01335},
  doi          = {10.1109/CVPR52688.2022.01335},
  bibsource    = {dblp computer science bibliography, https://dblp.org}
}

@inproceedings{Zhang2023ContinualLO,
  author       = {Peiyan Zhang and
                  Yuchen Yan and
                  Chaozhuo Li and
                  Senzhang Wang and
                  Xing Xie and
                  Guojie Song and
                  Sunghun Kim},
  editor       = {Hsin{-}Hsi Chen and
                  Wei{-}Jou (Edward) Duh and
                  Hen{-}Hsen Huang and
                  Makoto P. Kato and
                  Josiane Mothe and
                  Barbara Poblete},
  title        = {Continual Learning on Dynamic Graphs via Parameter Isolation},
  booktitle    = {Proceedings of the 46th International {ACM} {SIGIR} Conference on
                  Research and Development in Information Retrieval, {SIGIR} 2023, Taipei,
                  Taiwan, July 23-27, 2023},
  pages        = {601--611},
  publisher    = {{ACM}},
  year         = {2023},
  url          = {https://doi.org/10.1145/3539618.3591652},
  doi          = {10.1145/3539618.3591652},
  bibsource    = {dblp computer science bibliography, https://dblp.org}
}

@article{Zhang2021HierarchicalPN,
  author       = {Xikun Zhang and
                  Dongjin Song and
                  Dacheng Tao},
  title        = {Hierarchical Prototype Networks for Continual Graph Representation
                  Learning},
  journal      = {{IEEE} Trans. Pattern Anal. Mach. Intell.},
  volume       = {45},
  number       = {4},
  pages        = {4622--4636},
  year         = {2023},
  url          = {https://doi.org/10.1109/TPAMI.2022.3186909},
  doi          = {10.1109/TPAMI.2022.3186909},
  bibsource    = {dblp computer science bibliography, https://dblp.org}
}

@inproceedings{Gao2024BeyondPL,
  author       = {Xinyuan Gao and
                  Songlin Dong and
                  Yuhang He and
                  Qiang Wang and
                  Yihong Gong},
  editor       = {Ales Leonardis and
                  Elisa Ricci and
                  Stefan Roth and
                  Olga Russakovsky and
                  Torsten Sattler and
                  G{\"{u}}l Varol},
  title        = {Beyond Prompt Learning: Continual Adapter for Efficient Rehearsal-Free
                  Continual Learning},
  booktitle    = {Computer Vision - {ECCV} 2024 - 18th European Conference, Milan, Italy,
                  September 29-October 4, 2024, Proceedings, Part {LXXXV}},
  series       = {Lecture Notes in Computer Science},
  volume       = {15143},
  pages        = {89--106},
  publisher    = {Springer},
  year         = {2024},
  url          = {https://doi.org/10.1007/978-3-031-73013-9\_6},
  doi          = {10.1007/978-3-031-73013-9\_6},
  bibsource    = {dblp computer science bibliography, https://dblp.org}
}

@inproceedings{Fang2022UniversalPT,
  author       = {Taoran Fang and
                  Yunchao Zhang and
                  Yang Yang and
                  Chunping Wang and
                  Lei Chen},
  editor       = {Alice Oh and
                  Tristan Naumann and
                  Amir Globerson and
                  Kate Saenko and
                  Moritz Hardt and
                  Sergey Levine},
  title        = {Universal Prompt Tuning for Graph Neural Networks},
  booktitle    = {Advances in Neural Information Processing Systems 36: Annual Conference
                  on Neural Information Processing Systems 2023, NeurIPS 2023, New Orleans,
                  LA, USA, December 10 - 16, 2023},
  year         = {2023},
  url          = {http://papers.nips.cc/paper\_files/paper/2023/hash/a4a1ee071ce0fe63b83bce507c9dc4d7-Abstract-Conference.html},
  bibsource    = {dblp computer science bibliography, https://dblp.org}
}

@inproceedings{Liu2023GraphPromptUP,
  author       = {Zemin Liu and
                  Xingtong Yu and
                  Yuan Fang and
                  Xinming Zhang},
  editor       = {Ying Ding and
                  Jie Tang and
                  Juan F. Sequeda and
                  Lora Aroyo and
                  Carlos Castillo and
                  Geert{-}Jan Houben},
  title        = {GraphPrompt: Unifying Pre-Training and Downstream Tasks for Graph
                  Neural Networks},
  booktitle    = {Proceedings of the {ACM} Web Conference 2023, {WWW} 2023, Austin,
                  TX, USA, 30 April 2023 - 4 May 2023},
  pages        = {417--428},
  publisher    = {{ACM}},
  year         = {2023},
  url          = {https://doi.org/10.1145/3543507.3583386},
  doi          = {10.1145/3543507.3583386},
  bibsource    = {dblp computer science bibliography, https://dblp.org}
}

@article{Wang2025PromptDrivenCG,
  author       = {Qi Wang and
                  Tianfei Zhou and
                  Ye Yuan and
                  Rui Mao},
  title        = {Prompt-Driven Continual Graph Learning},
  journal      = {CoRR},
  volume       = {abs/2502.06327},
  year         = {2025},
  url          = {https://doi.org/10.48550/arXiv.2502.06327},
  doi          = {10.48550/ARXIV.2502.06327},
  eprinttype    = {arXiv},
  eprint       = {2502.06327},
  bibsource    = {dblp computer science bibliography, https://dblp.org}
}

@inproceedings{niu2024replayandforgetfree,
  author       = {Chaoxi Niu and
                  Guansong Pang and
                  Ling Chen and
                  Bing Liu},
  editor       = {Amir Globersons and
                  Lester Mackey and
                  Danielle Belgrave and
                  Angela Fan and
                  Ulrich Paquet and
                  Jakub M. Tomczak and
                  Cheng Zhang},
  title        = {Replay-and-Forget-Free Graph Class-Incremental Learning: {A} Task
                  Profiling and Prompting Approach},
  booktitle    = {Advances in Neural Information Processing Systems 38: Annual Conference
                  on Neural Information Processing Systems 2024, NeurIPS 2024, Vancouver,
                  BC, Canada, December 10 - 15, 2024},
  year         = {2024},
  url          = {http://papers.nips.cc/paper\_files/paper/2024/hash/a07e87ecfa8a651d62257571669b0150-Abstract-Conference.html},
  bibsource    = {dblp computer science bibliography, https://dblp.org}
}

@inproceedings{Liu2023GapformerGT,
  author       = {Chuang Liu and
                  Yibing Zhan and
                  Xueqi Ma and
                  Liang Ding and
                  Dapeng Tao and
                  Jia Wu and
                  Wenbin Hu},
  title        = {Gapformer: Graph Transformer with Graph Pooling for Node Classification},
  booktitle    = {Proceedings of the Thirty-Second International Joint Conference on
                  Artificial Intelligence, {IJCAI} 2023, 19th-25th August 2023, Macao,
                  SAR, China},
  pages        = {2196--2205},
  publisher    = {ijcai.org},
  year         = {2023},
  url          = {https://doi.org/10.24963/ijcai.2023/244},
  doi          = {10.24963/IJCAI.2023/244},
  bibsource    = {dblp computer science bibliography, https://dblp.org}
}

@article{Hu2020OpenGB,
  title={Open Graph Benchmark: Datasets for Machine Learning on Graphs},
  author={Weihua Hu and Matthias Fey and Marinka Zitnik and Yuxiao Dong and Hongyu Ren and Bowen Liu and Michele Catasta and Jure Leskovec},
  journal={ArXiv},
  year={2020},
  volume={abs/2005.00687},
  url={https://api.semanticscholar.org/CorpusID:218487328}
}

@inproceedings{Tang2023GraphGPTGI,
  author       = {Jiabin Tang and
                  Yuhao Yang and
                  Wei Wei and
                  Lei Shi and
                  Lixin Su and
                  Suqi Cheng and
                  Dawei Yin and
                  Chao Huang},
  editor       = {Grace Hui Yang and
                  Hongning Wang and
                  Sam Han and
                  Claudia Hauff and
                  Guido Zuccon and
                  Yi Zhang},
  title        = {GraphGPT: Graph Instruction Tuning for Large Language Models},
  booktitle    = {Proceedings of the 47th International {ACM} {SIGIR} Conference on
                  Research and Development in Information Retrieval, {SIGIR} 2024, Washington
                  DC, USA, July 14-18, 2024},
  pages        = {491--500},
  publisher    = {{ACM}},
  year         = {2024},
  url          = {https://doi.org/10.1145/3626772.3657775},
  doi          = {10.1145/3626772.3657775},
  bibsource    = {dblp computer science bibliography, https://dblp.org}
}

@misc{liu2021pretrainpromptpredictsystematic,
      title={Pre-train, Prompt, and Predict: A Systematic Survey of Prompting Methods in Natural Language Processing}, 
      author={Pengfei Liu and Weizhe Yuan and Jinlan Fu and Zhengbao Jiang and Hiroaki Hayashi and Graham Neubig},
      year={2021},
      eprint={2107.13586},
      archivePrefix={arXiv},
      primaryClass={cs.CL},
      url={https://arxiv.org/abs/2107.13586}, 
}

@inproceedings{Sinha2015AnOO,
  author       = {Arnab Sinha and
                  Zhihong Shen and
                  Yang Song and
                  Hao Ma and
                  Darrin Eide and
                  Bo{-}June Paul Hsu and
                  Kuansan Wang},
  editor       = {Aldo Gangemi and
                  Stefano Leonardi and
                  Alessandro Panconesi},
  title        = {An Overview of Microsoft Academic Service {(MAS)} and Applications},
  booktitle    = {Proceedings of the 24th International Conference on World Wide Web
                  Companion, {WWW} 2015, Florence, Italy, May 18-22, 2015 - Companion
                  Volume},
  pages        = {243--246},
  publisher    = {{ACM}},
  year         = {2015},
  url          = {https://doi.org/10.1145/2740908.2742839},
  doi          = {10.1145/2740908.2742839},
  bibsource    = {dblp computer science bibliography, https://dblp.org}
}

@misc{houlsby2019parameterefficienttransferlearningnlp,
      title={Parameter-Efficient Transfer Learning for NLP}, 
      author={Neil Houlsby and Andrei Giurgiu and Stanislaw Jastrzebski and Bruna Morrone and Quentin de Laroussilhe and Andrea Gesmundo and Mona Attariyan and Sylvain Gelly},
      year={2019},
      eprint={1902.00751},
      archivePrefix={arXiv},
      primaryClass={cs.LG},
      url={https://arxiv.org/abs/1902.00751}, 
}

@article{liu2023CaT,
    author={Yilun Liu and Ruihong Qiu and Zi Huang},
    title={CaT: Balanced Continual Graph Learning with  Graph Condensation},
    journal={Arxiv},
    year={2023},
    volume={abs/2309.09455},
    url={https://arxiv.org/abs/2309.09455}
}

@inproceedings{Zhang2022SparsifiedSubgraphMemoryforCGRL,
  author={Xikun Zhang and Dongjin Song and Dacheng Tao},
  booktitle={2022 IEEE International Conference on Data Mining (ICDM)}, 
  title={Sparsified Subgraph Memory for Continual Graph Representation Learning}, 
  year={2022},
  volume={},
  number={},
  pages={1335-1340},
  doi={10.1109/ICDM54844.2022.00177}
}

@article{Liu2024PUMA,
    author={Yilun Liu and Ruihong Qiu and Yanran Tang and Hongzhi Yin and Zi Huang},
    title={PUMA: Efficient Continual Graph Learning for Node Classification with Graph Condensation},
    journal={Arxiv},
    year={2024},
    volume={abs/2312.14439},
    url={http://arxiv.org/abs/2312.14439}
}

@inproceedings{Zhou2021Overcomingcatastrophicforgetting,
  author       = {Fan Zhou and
                  Chengtai Cao},
  title        = {Overcoming Catastrophic Forgetting in Graph Neural Networks with Experience
                  Replay},
  booktitle    = {Thirty-Fifth {AAAI} Conference on Artificial Intelligence, {AAAI}
                  2021, Thirty-Third Conference on Innovative Applications of Artificial
                  Intelligence, {IAAI} 2021, The Eleventh Symposium on Educational Advances
                  in Artificial Intelligence, {EAAI} 2021, Virtual Event, February 2-9,
                  2021},
  pages        = {4714--4722},
  publisher    = {{AAAI} Press},
  year         = {2021},
  url          = {https://doi.org/10.1609/aaai.v35i5.16602},
  doi          = {10.1609/AAAI.V35I5.16602},
  bibsource    = {dblp computer science bibliography, https://dblp.org}
}

@inproceedings{mas,
  author       = {Rahaf Aljundi and
                  Francesca Babiloni and
                  Mohamed Elhoseiny and
                  Marcus Rohrbach and
                  Tinne Tuytelaars},
  editor       = {Vittorio Ferrari and
                  Martial Hebert and
                  Cristian Sminchisescu and
                  Yair Weiss},
  title        = {Memory Aware Synapses: Learning What (not) to Forget},
  booktitle    = {Computer Vision - {ECCV} 2018 - 15th European Conference, Munich,
                  Germany, September 8-14, 2018, Proceedings, Part {III}},
  series       = {Lecture Notes in Computer Science},
  volume       = {11207},
  pages        = {144--161},
  publisher    = {Springer},
  year         = {2018},
  url          = {https://doi.org/10.1007/978-3-030-01219-9\_9},
  doi          = {10.1007/978-3-030-01219-9\_9},
  bibsource    = {dblp computer science bibliography, https://dblp.org}
}

@inproceedings{TWP,
  author       = {Huihui Liu and
                  Yiding Yang and
                  Xinchao Wang},
  title        = {Overcoming Catastrophic Forgetting in Graph Neural Networks},
  booktitle    = {Thirty-Fifth {AAAI} Conference on Artificial Intelligence, {AAAI}
                  2021, Thirty-Third Conference on Innovative Applications of Artificial
                  Intelligence, {IAAI} 2021, The Eleventh Symposium on Educational Advances
                  in Artificial Intelligence, {EAAI} 2021, Virtual Event, February 2-9,
                  2021},
  pages        = {8653--8661},
  publisher    = {{AAAI} Press},
  year         = {2021},
  url          = {https://doi.org/10.1609/aaai.v35i10.17049},
  doi          = {10.1609/AAAI.V35I10.17049},
  bibsource    = {dblp computer science bibliography, https://dblp.org}
}

@inproceedings{Lwf,
  author       = {Zhizhong Li and
                  Derek Hoiem},
  editor       = {Bastian Leibe and
                  Jiri Matas and
                  Nicu Sebe and
                  Max Welling},
  title        = {Learning Without Forgetting},
  booktitle    = {Computer Vision - {ECCV} 2016 - 14th European Conference, Amsterdam,
                  The Netherlands, October 11-14, 2016, Proceedings, Part {IV}},
  series       = {Lecture Notes in Computer Science},
  volume       = {9908},
  pages        = {614--629},
  publisher    = {Springer},
  year         = {2016},
  url          = {https://doi.org/10.1007/978-3-319-46493-0\_37},
  doi          = {10.1007/978-3-319-46493-0\_37},
  bibsource    = {dblp computer science bibliography, https://dblp.org}
}

@inproceedings{SSM,
  author       = {Xikun Zhang and
                  Dongjin Song and
                  Dacheng Tao},
  editor       = {Xingquan Zhu and
                  Sanjay Ranka and
                  My T. Thai and
                  Takashi Washio and
                  Xindong Wu},
  title        = {Sparsified Subgraph Memory for Continual Graph Representation Learning},
  booktitle    = {{IEEE} International Conference on Data Mining, {ICDM} 2022, Orlando,
                  FL, USA, November 28 - Dec. 1, 2022},
  pages        = {1335--1340},
  publisher    = {{IEEE}},
  year         = {2022},
  url          = {https://doi.org/10.1109/ICDM54844.2022.00177},
  doi          = {10.1109/ICDM54844.2022.00177},
  bibsource    = {dblp computer science bibliography, https://dblp.org}
}

@inproceedings{ERGNN,
  author       = {Fan Zhou and
                  Chengtai Cao},
  title        = {Overcoming Catastrophic Forgetting in Graph Neural Networks with Experience
                  Replay},
  booktitle    = {Thirty-Fifth {AAAI} Conference on Artificial Intelligence, {AAAI}
                  2021, Thirty-Third Conference on Innovative Applications of Artificial
                  Intelligence, {IAAI} 2021, The Eleventh Symposium on Educational Advances
                  in Artificial Intelligence, {EAAI} 2021, Virtual Event, February 2-9,
                  2021},
  pages        = {4714--4722},
  publisher    = {{AAAI} Press},
  year         = {2021},
  url          = {https://doi.org/10.1609/aaai.v35i5.16602},
  doi          = {10.1609/AAAI.V35I5.16602},
  bibsource    = {dblp computer science bibliography, https://dblp.org}
}

@inproceedings{Zhang2024Topology-awareembeddingmemoryforcl,
  author       = {Xikun Zhang and
                  Dongjin Song and
                  Yixin Chen and
                  Dacheng Tao},
  editor       = {Ricardo Baeza{-}Yates and
                  Francesco Bonchi},
  title        = {Topology-aware Embedding Memory for Continual Learning on Expanding
                  Networks},
  booktitle    = {Proceedings of the 30th {ACM} {SIGKDD} Conference on Knowledge Discovery
                  and Data Mining, {KDD} 2024, Barcelona, Spain, August 25-29, 2024},
  pages        = {4326--4337},
  publisher    = {{ACM}},
  year         = {2024},
  url          = {https://doi.org/10.1145/3637528.3671732},
  doi          = {10.1145/3637528.3671732},
  bibsource    = {dblp computer science bibliography, https://dblp.org}
}

@inproceedings{gcn,
  author       = {Thomas N. Kipf and
                  Max Welling},
  title        = {Semi-Supervised Classification with Graph Convolutional Networks},
  booktitle    = {5th International Conference on Learning Representations, {ICLR} 2017,
                  Toulon, France, April 24-26, 2017, Conference Track Proceedings},
  publisher    = {OpenReview.net},
  year         = {2017},
  url          = {https://openreview.net/forum?id=SJU4ayYgl},
  bibsource    = {dblp computer science bibliography, https://dblp.org}
}

@inproceedings{Wu2019SimplifyingGC,
  author       = {Felix Wu and
                  Amauri H. Souza Jr. and
                  Tianyi Zhang and
                  Christopher Fifty and
                  Tao Yu and
                  Kilian Q. Weinberger},
  editor       = {Kamalika Chaudhuri and
                  Ruslan Salakhutdinov},
  title        = {Simplifying Graph Convolutional Networks},
  booktitle    = {Proceedings of the 36th International Conference on Machine Learning,
                  {ICML} 2019, 9-15 June 2019, Long Beach, California, {USA}},
  series       = {Proceedings of Machine Learning Research},
  volume       = {97},
  pages        = {6861--6871},
  publisher    = {{PMLR}},
  year         = {2019},
  url          = {http://proceedings.mlr.press/v97/wu19e.html},
  bibsource    = {dblp computer science bibliography, https://dblp.org}
}

@inproceedings{Klicpera2018PredictTP,
  author       = {Johannes Klicpera and
                  Aleksandar Bojchevski and
                  Stephan G{\"{u}}nnemann},
  title        = {Predict then Propagate: Graph Neural Networks meet Personalized PageRank},
  booktitle    = {7th International Conference on Learning Representations, {ICLR} 2019,
                  New Orleans, LA, USA, May 6-9, 2019},
  publisher    = {OpenReview.net},
  year         = {2019},
  url          = {https://openreview.net/forum?id=H1gL-2A9Ym},
  bibsource    = {dblp computer science bibliography, https://dblp.org}
}

@inproceedings{Brockschmidt2019GNNFiLMGN,
  author       = {Marc Brockschmidt},
  title        = {GNN-FiLM: Graph Neural Networks with Feature-wise Linear Modulation},
  booktitle    = {Proceedings of the 37th International Conference on Machine Learning,
                  {ICML} 2020, 13-18 July 2020, Virtual Event},
  series       = {Proceedings of Machine Learning Research},
  volume       = {119},
  pages        = {1144--1152},
  publisher    = {{PMLR}},
  year         = {2020},
  url          = {http://proceedings.mlr.press/v119/brockschmidt20a.html},
  bibsource    = {dblp computer science bibliography, https://dblp.org}
}

@article{Rossi2020SIGNSI,
  title={SIGN: Scalable Inception Graph Neural Networks},
  author={Emanuele Rossi and Fabrizio Frasca and Benjamin Paul Chamberlain and Davide Eynard and Michael M. Bronstein and Federico Monti},
  journal={ArXiv},
  year={2020},
  volume={abs/2004.11198},
  url={https://api.semanticscholar.org/CorpusID:216080440}
}

@inproceedings{Perez2017FiLMVR,
  author       = {Ethan Perez and
                  Florian Strub and
                  Harm de Vries and
                  Vincent Dumoulin and
                  Aaron C. Courville},
  editor       = {Sheila A. McIlraith and
                  Kilian Q. Weinberger},
  title        = {FiLM: Visual Reasoning with a General Conditioning Layer},
  booktitle    = {Proceedings of the Thirty-Second {AAAI} Conference on Artificial Intelligence,
                  (AAAI-18), the 30th innovative Applications of Artificial Intelligence
                  (IAAI-18), and the 8th {AAAI} Symposium on Educational Advances in
                  Artificial Intelligence (EAAI-18), New Orleans, Louisiana, USA, February
                  2-7, 2018},
  pages        = {3942--3951},
  publisher    = {{AAAI} Press},
  year         = {2018},
  url          = {https://doi.org/10.1609/aaai.v32i1.11671},
  doi          = {10.1609/AAAI.V32I1.11671},
  bibsource    = {dblp computer science bibliography, https://dblp.org}
}

@inproceedings{Aozm,
author = {McAuley, Julian and Targett, Christopher and Shi, Qinfeng and van den Hengel, Anton},
title = {Image-Based Recommendations on Styles and Substitutes},
year = {2015},
isbn = {9781450336215},
publisher = {Association for Computing Machinery},
address = {New York, NY, USA},
url = {https://doi.org/10.1145/2766462.2767755},
doi = {10.1145/2766462.2767755},
booktitle = {Proceedings of the 38th International ACM SIGIR Conference on Research and Development in Information Retrieval},
pages = {43–52},
numpages = {10},
location = {Santiago, Chile},
series = {SIGIR '15}
}

@inproceedings{liu2023one,
  author       = {Hao Liu and
                  Jiarui Feng and
                  Lecheng Kong and
                  Ningyue Liang and
                  Dacheng Tao and
                  Yixin Chen and
                  Muhan Zhang},
  title        = {One For All: Towards Training One Graph Model For All Classification
                  Tasks},
  booktitle    = {The Twelfth International Conference on Learning Representations,
                  {ICLR} 2024, Vienna, Austria, May 7-11, 2024},
  publisher    = {OpenReview.net},
  year         = {2024},
  url          = {https://openreview.net/forum?id=4IT2pgc9v6},
  bibsource    = {dblp computer science bibliography, https://dblp.org}
}

@inproceedings{DeLoMe,
  author       = {Chaoxi Niu and
                  Guansong Pang and
                  Ling Chen},
  editor       = {Ulle Endriss and
                  Francisco S. Melo and
                  Kerstin Bach and
                  Alberto Jos{\'{e}} Bugar{\'{\i}}n Diz and
                  Jose Maria Alonso{-}Moral and
                  Sen{\'{e}}n Barro and
                  Fredrik Heintz},
  title        = {Graph Continual Learning with Debiased Lossless Memory Replay},
  booktitle    = {{ECAI} 2024 - 27th European Conference on Artificial Intelligence,
                  19-24 October 2024, Santiago de Compostela, Spain - Including 13th
                  Conference on Prestigious Applications of Intelligent Systems {(PAIS}
                  2024)},
  series       = {Frontiers in Artificial Intelligence and Applications},
  volume       = {392},
  pages        = {1808--1815},
  publisher    = {{IOS} Press},
  year         = {2024},
  url          = {https://doi.org/10.3233/FAIA240692},
  doi          = {10.3233/FAIA240692},
  bibsource    = {dblp computer science bibliography, https://dblp.org}
}

\clearpage
\onecolumn
\appendix

\section{Theoretical Analysis of AMP}
\label{sec:theoretical_analysis}

In this section, we provide a theoretical analysis to ground our choice of Anchored Multi-hop Propagation (AMP) for task prototype generation. We aim to demonstrate that AMP is inherently designed to produce task prototypes with high inter-task separability and low intra-task variance. This property is crucial for minimizing the task classification error during inference, which is the cornerstone of our model's stability.

\subsection{Objective Formulation}
Let us consider two distinct tasks, $\tau_k$ and $\tau_j$, with their respective training node sets $\mathcal{V}_k^{\text{train}}$ and $\mathcal{V}_j^{\text{train}}$, and initial feature matrices $\mathbf{H}_k^{(0)}$ and $\mathbf{H}_j^{(0)}$. Our goal is to generate prototypes $\mathbf{p}_k$ and $\mathbf{p}_j$ that are maximally separable. An ideal prototype generation function, $f(\cdot)$, should achieve two objectives:

\begin{enumerate}
    \item \textbf{Minimize Intra-Task Variance:} The features of nodes within the same task should be compactly clustered around their task's prototype. We can define the intra-task variance for task $\tau_k$ as:
    \begin{equation}
        V_{\text{intra}}(\tau_k) = \mathbb{E}_{v \in \mathcal{V}_k^{\text{train}}} \left[ || f(\mathbf{H}^{(0)})_v - \mathbf{p}_k ||^2 \right]
    \end{equation}
    where $f(\mathbf{H}^{(0)})_v$ is the transformed feature of node $v$.

    \item \textbf{Maximize Inter-Task Distance:} The prototypes of different tasks should be far apart from each other in the embedding space. We define the squared inter-task distance as:
    \begin{equation}
        D_{\text{inter}}(\tau_k, \tau_j)^2 = || \mathbf{p}_k - \mathbf{p}_j ||^2
    \end{equation}
\end{enumerate}
We will now analyze how each component of AMP contributes to optimizing these two objectives.

\subsection{Intra-Task Variance Reduction via Graph Diffusion}
The core of AMP is the APPNP propagation rule (Eq. (8)). Let's analyze the iterative diffusion component, $\mathbf{Z}^{(i+1)} \propto \mathbf{S}\mathbf{Z}^{(i)}$. This operation is a form of graph Laplacian smoothing. In spectral graph theory, multiplying features by the normalized adjacency matrix $\mathbf{S}$ is equivalent to applying a low-pass filter to the graph signal (i.e., the node features).

A low-pass filter attenuates high-frequency components of the signal. In the context of graphs, high-frequency components correspond to large feature differences between adjacent nodes. Therefore, the diffusion process forces the features of connected nodes to become more similar. This process is equivalent to minimizing the graph's Dirichlet energy:
\begin{equation}
    \mathcal{E}(\mathbf{Z}) = \frac{1}{2} \sum_{u,v} A_{uv} \left|\left| \frac{\mathbf{z}_u}{\sqrt{d_u}} - \frac{\mathbf{z}_v}{\sqrt{d_v}} \right|\right|^2 = \text{Tr}(\mathbf{Z}^\top (\mathbf{I}-\mathbf{S})\mathbf{Z})
\end{equation}
where $\text{Tr}(\cdot)$ is the trace of a matrix. Iteratively applying $\mathbf{S}$ minimizes this energy, effectively smoothing the features across the graph.

Our key assumption, standard in graph-based learning, is that \textit{intra-task connectivity is significantly denser than inter-task connectivity}. Under this assumption, the feature smoothing will primarily occur *within* the node set of a given task $\tau_k$. Consequently, the features of nodes in $\mathcal{V}_k^{\text{train}}$ will converge towards a common representation, leading to a direct reduction in the intra-task variance, $V_{\text{intra}}(\tau_k)$. The resulting prototype $\mathbf{p}_k$, being the mean of these smoothed features, becomes a more stable and robust representation of the entire task.

\subsection{Preserving Inter-Task Separability}
While feature smoothing is beneficial for reducing intra-task variance, excessive smoothing (oversmoothing) is detrimental, as it would cause all nodes in the graph to converge to the same vector, collapsing all prototypes and making $D_{\text{inter}} \to 0$. AMP employs two mechanisms to counteract this and preserve separability.

\textbf{1. Anchoring with Teleport Probability:} The teleport term $\alpha\mathbf{H}^{(0)}$ in Eq. (8) acts as a crucial regularizer. At each propagation step, it pulls the smoothed features back towards their initial, task-specific representations. This "anchoring" ensures that the feature distributions of different tasks, which are assumed to be initially distinct, do not completely merge. It preserves the global, task-level differences while the diffusion term smooths out local, intra-task variations.

\textbf{2. Multi-scale Feature Concatenation:} Different tasks may be characterized by different topological scales. For instance, one task's graph may consist of many small, tight clusters, while another's may be defined by long-range dependencies. A prototype based on a single propagation depth $K$ might fail to capture these diverse structural fingerprints. By concatenating features from multiple hops $\{h_1, \dots, h_m\}$ (Eq. (9)), $\mathbf{H}_{\text{AMP}}$ creates a far richer and more discriminative embedding. This allows the prototype to simultaneously encode local patterns (from small $h_i$) and global community structure (from large $h_i$), significantly increasing the likelihood that the prototypes for structurally or featurally distinct tasks will be distant from each other, thus maximizing $D_{\text{inter}}$.

\subsection{Minimizing Task Classification Error}
The task-ID inference is a nearest-centroid classification problem based on cosine similarity, which is closely related to Euclidean distance for normalized vectors. The probability of misclassifying a test prototype $\mathbf{p}_{\text{test}}$ belonging to task $\tau_k$ as belonging to task $\tau_j$ is a function of the distance to their respective centroids and the variance of the features. For a simplified two-class case, the Bayes classification error rate $P_e$ is bounded by a function that decreases as the ratio of inter-task distance to intra-task variance increases:
\begin{equation}
    P_e(\tau_k \to \tau_j) \le \exp\left(-\frac{D_{\text{inter}}(\tau_k, \tau_j)^2}{8 \cdot \max(V_{\text{intra}}(\tau_k), V_{\text{intra}}(\tau_j))}\right)
\end{equation}
From our analysis, AMP is explicitly designed to optimize this ratio. The graph diffusion component minimizes the denominator ($V_{\text{intra}}$), while the anchoring and multi-scale concatenation mechanisms work to preserve or maximize the numerator ($D_{\text{inter}}$). By generating prototypes with these desirable geometric properties, AMP provides a strong foundation for a classifier that can robustly distinguish between tasks, thereby minimizing the task-ID prediction error and ensuring the overall stability of the continual learning system.

\paragraph{Datasets and Baselines.}
We perform evaluations on eight benchmark datasets, which span several domains. Specifically, these include three citation networks: CoraFull~\cite{McCallum2000AutomatingTC}, Arxiv~\cite{Hu2020OpenGB}, and Citeseer~\cite{Yang2016RevisitingSL}; one social network: Reddit~\cite{Hamilton2017InductiveRL}; three product co-purchase networks: Products~\cite{Hu2020OpenGB}, Photo~\cite{Aozm}, and Computer~\cite{Aozm}; and one webpage-based network: WikiCS~\cite{liu2023one}, with statistics detailed in Table~\ref{tab:dataset_stats}. Our baselines cover the spectrum of GCL strategies,including regularization, replay-base, and parameter isolation leading methods. We use Joint training and Oracle models as upper bounds.

\section{Details of Datesets}

\begin{itemize}

    \item \textbf{Citeseer}\cite{Yang2016RevisitingSL}: Another widely-used citation network where nodes are scientific publications and edges are citations. The papers are classified into six categories. Similar to Cora, node features are based on the document's textual content.

    \item \textbf{CoraFull}\cite{McCallum2000AutomatingTC}: A larger citation network encompassing 70 classes. Nodes correspond to academic papers and edges denote citation links between them.

    \item \textbf{Arxiv}\cite{Hu2020OpenGB}: This dataset is constructed from the collaboration network of Computer Science (CS) papers on the arXiv platform, indexed by MAG \cite{Sinha2015AnOO}. Nodes, which represent CS papers, are classified into 40 subject areas. Node features are generated by averaging the word embeddings of the titles and abstracts.

    \item \textbf{Reddit}\cite{Hamilton2017InductiveRL}: A social network dataset derived from Reddit posts made in September 2014. Here, nodes are individual posts, and an edge connects two posts if the same user has commented on both. The objective is to classify posts into their corresponding communities (subreddits). Node features are engineered from post attributes such as title, content, score, and comment count.

    \item \textbf{Products}\cite{Hu2020OpenGB}: An Amazon product co-purchase network where nodes represent products and edges indicate that two products are frequently bought together. Node features are derived from dimensionality-reduced bag-of-words representations of the product descriptions.
    
    \item \textbf{WikiCS}\cite{liu2023one}: A dataset constructed based on computer science articles from Wikipedia. Nodes represent articles, and edges are hyperlinks between them. The task is to classify the articles into 10 categories based on their textual content.
    
    \item \textbf{Computer}\cite{Aozm}: A dataset from Amazon's group-buying network. Nodes represent goods related to computers, while edges indicate the relationship where goods are purchased together. The task is to classify them into their respective categories. The node features are derived from product reviews.

    \item \textbf{Photo}\cite{Aozm}: Similar to the Computer dataset, this is also part of Amazon's group-buying network. Nodes are goods related to photography (such as cameras and lenses), while edges connect goods that are usually purchased together. Node features are generated by the bag-of-words representation of product reviews.

\end{itemize}

\section{Details of baselines}
We evaluate our method on a suite of leading CGL methods. including three mainstream categories based on Replay, regularization, and parameter isolation.

\begin{itemize}

    \item \textbf{Finetuning}\cite{gcn}is denotes the backbone GNN without continual learning technique. Therefore,this can be viewed as the lower bound on the continual learning performance.
        
    \item \textbf{LwF}\cite{LwF} employs knowledge distillation to maintain performance on old tasks. It minimizes the discrepancy between the output logits of the previous and current models, effectively transferring knowledge without retaining old data.

    \item\textbf{MAS}\cite{mas} also preserves vital parameters from past tasks, but bases the importance on the sensitivity of the model's predictions to parameter shifts.

    \item\textbf{TWP} \cite{TWP} focuses on preserving important parameters within a topological aggregation framework and introduces a loss minimization strategy for previous tasks that relies on these preserved parameters.

    \item\textbf{ER-GNN} \cite{ERGNN} is a replay-based method that constructs a memory buffer by storing representative nodes selected from previous tasks.

    \item\textbf{SSM} \cite{SSM} enhances graph continual learning by explicitly incorporating topological information. It stores sparsified computation subgraphs of selected nodes in its memory.
    
    \item\textbf{PDGNNs} \cite{Zhang2024Topology-awareembeddingmemoryforcl} decouple trainable parameters from computational subgraphs through topology-aware embeddings (TEs). These TEs are compressed representations of ego-subnetworks that significantly reduce memory overhead.
    
    \item\textbf{DeLoMe} \cite{DeLoMe}  constructed a memory composed of lossless node prototypes, aiming to fully capture and store the graph information of old tasks. Meanwhile, to address the category imbalance that occurs when new and old data are mixed, this method further proposes a de-biased GCL loss function.
    
    \item\textbf{TPP} \cite{niu2024replayandforgetfree} a GCL method that avoids both data replay and catastrophic forgetting, leverages a Laplacian smoothing technique for precise task-ID inference and employs graph prompts to dynamically adapt the GNN into specialized modules for each task.
    
    \item\textbf{Joint} is base the backbone SGC can can get access to the data of all tasks, representing a theoretical upper bound of graph class-incremental learning
    
    \item\textbf{Oracle} is base the backbone SGC can can get access to the data of all tasksand task IDs, representing a stronger upper bound of graph class-incremental learning.
    
\end{itemize}

\begin{figure}[t]
    \centering
    \Description{}
    \includegraphics[width=1\columnwidth]{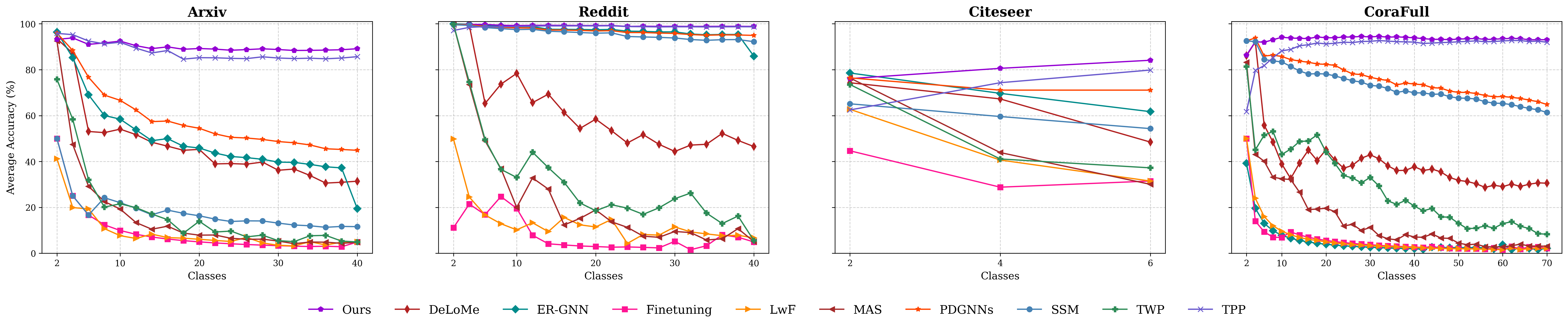}
    \caption{Performance comparison of our method, \textbf{TAAM} (Ours, purple line), against leading CGL methods on four datasetsin inductive learning scenario.}
    \label{fig:faa}
\end{figure}

\begin{figure}[t]
    \centering
    \Description{}
    \includegraphics[width=1\columnwidth]{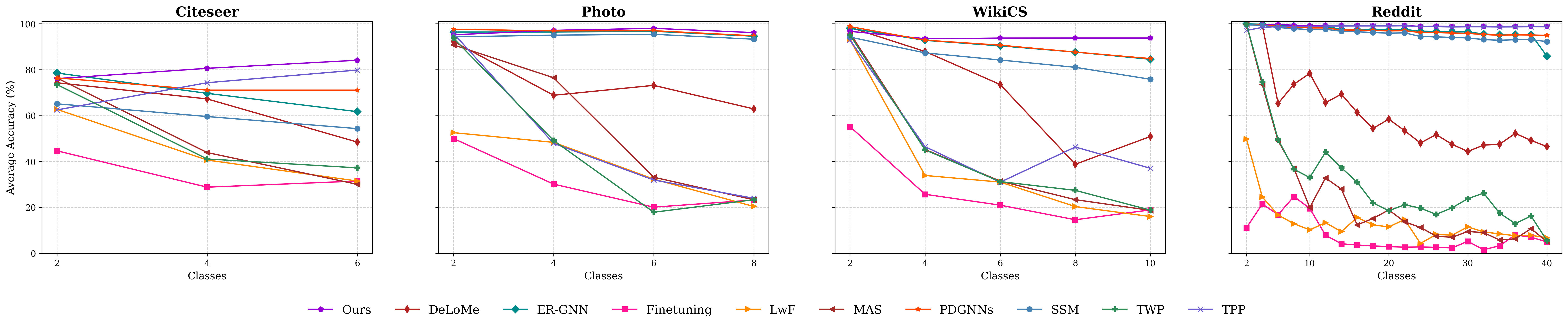}
    \caption{Performance comparison of our method, \textbf{TAAM} (Ours, purple line), against leading CGL methods on other four datasets in inductive learning scenario.}
    \label{fig:tradeoff}
\end{figure}

\section{Implementation Details}

All experiments are conducted on a server equipped with an Intel(R) Xeon(R) Platinum 8368Q CPU @ 2.60GHz, 256 GB of RAM, and a single NVIDIA L40 GPU. The software stack includes Python 3.8.16, PyTorch 1.10.1, DGL 0.9.0, OGB 1.3.6, and CUDA 12.4.

\section{Evaluation Metrics}

To comprehensively evaluate the model's performance in a continual learning setting, we adopt two standard metrics: Average Accuracy and Average Forgetting . Our evaluation protocol is based on the accuracy matrix $M \in \mathbb{R}^{T \times T}$, where $T$ represents the total number of tasks. An element $M_{t,j}$ of this matrix denotes the classification accuracy on task $j$ after the model has been sequentially trained up to task $t$.

\noindent\textbf{Average Accuracy (AA)} measures the overall performance of the model across all tasks after the entire training sequence is complete. It is calculated by averaging the accuracies on all tasks $j$ after the final task $T$ has been learned. A higher AA signifies a better-performing model. The formula is:
$$
\text{AA} = \frac{\sum_{j=1}^{T} M_{T,j}}{T}
$$

\noindent\textbf{Average Forgetting (AF)} quantifies how much the model forgets previously learned tasks as it acquires new knowledge. It is computed by averaging the difference between the peak accuracy on a task (i.e., immediately after it was learned) and its accuracy after training on a subsequent task, averaged over all past tasks. The formula is given by:
$$
\text{AF} = \frac{\sum_{j=1}^{T-1} (M_{T,j} - M_{j,j})}{T-1}
$$

%%%%%%%%%%%%%%%%%%%%%%%%%%%%%%%%%%%%%%%%%%%%%%%%%%%%%%%%%%%%%%%%%%%%%%%%
\end{document}